\documentclass[11pt]{article}
\usepackage{amsmath}
\usepackage{tabularx}
\usepackage{array}
\usepackage{booktabs}
\usepackage{placeins} %
\usepackage{threeparttable}
\usepackage{makecell}
\usepackage[ruled,vlined]{algorithm2e}
\usepackage[most]{tcolorbox}

\usepackage[preprint]{acl}

\usepackage{times}
\usepackage{latexsym}

\usepackage[T1]{fontenc}

\usepackage[utf8]{inputenc}
\usepackage{soul}
\usepackage{microtype}

\usepackage{inconsolata}

\usepackage{graphicx}

\definecolor{oscar_color}{HTML}{f77f00}

\definecolor{kh}{HTML}{168aff}

\definecolor{rs}{HTML}{CC0000}

\definecolor{nk}{HTML}{800080}

\title{ChronoCulture: Evaluating Cultural Awareness Across Time in Large Language Models}

\author{Rusali Saha$\quad$ Nitya Khurana$\quad$ Ruihong Huang$\quad$ Kuan-Hao Huang\vspace{0.1em}\\
  Texas A\&M University\vspace{0.1em}\\
  \texttt{\{rs0921,nityakhurana,huangrh,khhuang\}@tamu.edu}}

\begin{document}
\maketitle

\begin{abstract}
Cultural attitudes and social norms evolve over time, yet most cultural
benchmarks for large language models (LLMs) treat culture as a static
attribute of a country or region. We introduce ChronoCulture, a
benchmark of 10,360 temporal pairs built from longitudinal survey data
across 72 countries. Each pair contains the same survey question asked
in the same country in an older and a newer year, with full population
response distributions as ground truth. Across six LLMs, we find that
models can often predict a country's attitudes at one point in time but
do not reliably capture how those attitudes change. Their predictions
for different years tend to stay close together, failing to capture much of the
observed change. Specifying the intended year in the prompt also offers
little overall improvement in alignment with survey responses.
These findings suggest that models often present a largely static
picture of cultural attitudes, highlighting the need to evaluate
cultural awareness across time.
\end{abstract}

\section{Introduction}

Large language models (LLMs) are increasingly deployed in settings that require cultural awareness, from globalized communication and education to social science and anthropology research. This has motivated a growing line of benchmarks that test whether models can represent cultural views and social norms across diverse countries and communities \citep{myung2024blend, chiu-etal-2025-culturalbench, alwajih-etal-2025-palm, li2024cultureparkboostingcrossculturalunderstanding}. However, most existing benchmarks treat culture as a fixed rather than evolving characteristic of a country or community, thereby evaluating models against a single cultural snapshot.

\begin{figure}[t]
    \centering
    \includegraphics[ width=\linewidth]{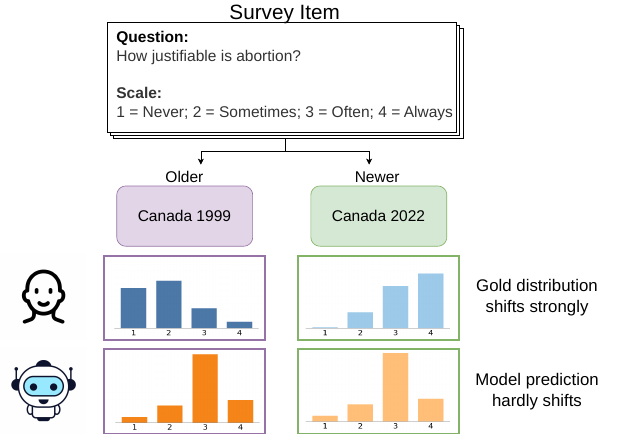}
    \caption{Limited temporal sensitivity: survey responses
    in Canada shift between 1999 and 2022, while Llama-3.3-70B predicts
    similar distributions for both years.}

    \label{fig:intro_fig}
\end{figure}

In reality, cultural attitudes and social norms are not static. They may evolve over time in response to economic development, generational change, and political transformation \citep{inglehart1997postmodernization, inglehart2018culture}. The extent of this change can also vary across societies: attitudes toward religion, gender roles, or trust may remain relatively stable in some countries while changing substantially in others \citep{inglehartbaker2000persistence}. 
This raises a central question: can language models represent how cultural attitudes evolve over time?

Figure~\ref{fig:intro_fig} illustrates a key motivation for our work.
Survey data show that attitudes toward abortion in Canada shifted markedly
between 1999 and 2022, yet the model produces similar predictions for both
periods. Evaluating a model against a single point in time provides only a partial view of its cultural awareness. A more complete evaluation requires comparing predictions for the same question and country across multiple time periods. 

To address this gap, we introduce \textbf{ChronoCulture}, a benchmark for
evaluating whether LLMs capture cultural change over time. ChronoCulture
is built from two long-running international survey programs, the World
Values Survey (WVS) \citep{haerpfer2022wvs} and the International
Social Survey Programme (ISSP) \citep{issp_programme}, covering 72 countries between 1985 and 2022. Its basic unit is a \emph{temporal pair}: the same survey question asked in the same country in an older and a newer year, with the full distribution of population responses
for each year as ground truth. We ask models to predict these distributions for each year and compare their predictions with the observed survey responses. 

Across six LLMs, we find that models can often predict a country's attitudes at one point in time, but do not reliably capture how they change. Their predictions for different years tend to stay close together despite observed change, and specifying the intended year in the prompt does not consistently improve alignment. These findings suggest that model predictions often present a static picture of cultural attitudes, rather than an account of how those attitudes evolve. Our contributions are as follows:
\begin{enumerate}
    \item We introduce ChronoCulture, a benchmark of 10,360 temporal
    pairs across 72 countries that evaluates whether LLMs capture
    cultural change, using full population response distributions rather
    than single answers.

    \item We develop an evaluation framework that assesses whether
    LLMs accurately predict both periods, capture cultural change
    and stability, and benefit from specifying the intended year.
    
    \item We release code to rebuild the benchmark and run all evaluations, for future research.\footnote{\href{https://github.com/lab-flair/ChronoCulture}{\nolinkurl{github.com/lab-flair/ChronoCulture}}. Due to licensing restrictions, we release scripts that rebuild the WVS and ISSP data from the original survey files.}
\end{enumerate}

\section{Related Works}
\paragraph{Socio-Cultural Knowledge Evaluation for Large Language Models}
Existing cultural benchmarks evaluate LLMs on everyday knowledge and commonsense \citep{myung2024blend,mrl-workshop-2025-global-piqa,singh-etal-2025-global}, cultural values and norms \citep{naous-etal-2024-beer,chiu-etal-2025-culturalbench,rao-etal-2025-normad,qiu-etal-2025-evaluating}, and multimodal cultural knowledge \citep{schneider-etal-2025-gimmick,li2026ravenea,tan2026blend,faraz2026indicvisionbench}. However, these benchmarks largely treat culture as a static attribute of a country, language, or region \citep{zhou2025culture,alkhamissi-etal-2026-hire}. We instead treat culture as evolving over time and examine whether LLMs capture changes in cultural attitudes.

\paragraph{Temporal-Aware Question Answering}
Temporal QA benchmarks primarily focus on changing factual knowledge, such as historical events and entity attributes \citep{chen2021a,tang-etal-2025-evowiki,qianyihu-etal-2025-time,piryani2025high,jiang-etal-2026-benchmarks}. Related vision-language research examines whether models can interpret
cultural content from different historical periods, including films,
artifacts, and cultural heritage images
\citep{jajoo2025cultureshift,ghaboura-etal-2025-time,ranjan2026culturalanachronismtemporalreasoning}.
We instead evaluate whether text-based LLMs capture changes in
population attitudes, using longitudinal survey responses.

\paragraph{Survey-based Opinion and Value Alignment}
OpinionQA, GlobalOpinionQA, and WorldValuesBench compare LLM outputs with survey-based human opinions across demographic groups and countries \citep{santurkar2023whose,durmus2024towards,zhao2024worldvaluesbench}. \citet{liu2026alignment} extend this comparison across WVS waves, finding
that LLMs align more with contemporary opinions. While these
works examine which population group or survey wave model responses
most closely resemble, we evaluate whether models capture how the same
population's attitudes change over time.

Concurrent with our work, \citet{daryani2026accuratespaceunreliabletime}
study cultural change using the Inglehart--Welzel map, which summarizes a
country's survey responses along two cultural dimensions. They find that
LLMs capture only part of the observed change, consistent with our
results. While they track changes in a country's overall cultural
position, ChronoCulture compares full response distributions for
individual survey questions within each country, across time. This
allows us to identify changes that may be hidden when multiple questions
are combined into an aggregate measure.

\section{Benchmark Creation}
We introduce ChronoCulture, a benchmark for evaluating whether LLMs can represent cultural attitudes at different time periods and capture how those attitudes change. It contains survey questions asked in the same countries at different time periods, so that each cultural topic can be examined over time within the same country.
 
We construct the benchmark using three design principles. First, we
ground cultural attitudes in people's survey responses for a given
country and year, drawing on surveys conducted across countries over multiple years. Second, we use full response distributions as ground truth rather than a single mean or majority answer,
preserving the diversity of opinions within each population. Third, we compare the same question within the same country across time, forming \emph{temporal pairs} of response distributions from two survey years.

The second design principle is particularly important for capturing cultural change. Several existing cultural benchmarks evaluate MCQs based on a single correct answer \citep{myung2024blend, romanou2025include, alwajih2025palmx, kim2024click, koto2024indoculture}. As we show in
Section~\ref{subsec:distribution evaluation} and
Figure~\ref{fig:mcq_limits}, a single correct answer can obscure how
cultural attitudes change.

\begin{figure}[t]
    \centering
    \includegraphics[width=\linewidth, trim=0 10 0 10, clip]{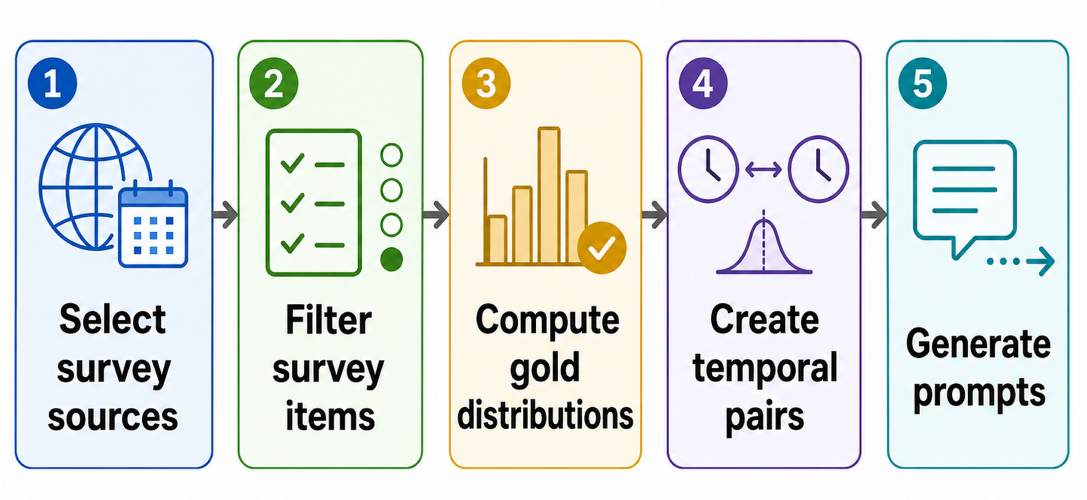}
    \caption{Overview: Benchmark construction pipeline.}
    \label{fig:pipeline_fig}
\end{figure}

The following subsections describe how we select and process survey data to construct temporal pairs. Figure~\ref{fig:pipeline_fig} summarizes this process.

\subsection{Data Sources}
We construct the benchmark from two international survey programs that collect longitudinal data on social attitudes and values
across countries: the World Values Survey (WVS) \citep{haerpfer2022wvs} and the International Social Survey Programme (ISSP) \citep{issp_programme}. Each record contains an individual's response to a question, with the country and survey year. We use WVS Waves 3--7 (1995--2022) and ISSP data spanning 1985--2022 (Appendix~\ref{app:data_sources}).

\subsection{Variable Selection}
We define a \emph{variable} as a survey question repeated in at least
two waves with a consistent, ordered response scale. Ordered options are
required by our evaluation method
(Section~\ref{subsec:temporal-pair-categorize}). Variables cover topics such as trust, religion, gender roles, and environmental attitudes. We exclude demographic metadata, behavioral frequency items, and country-specific questions that are not comparable across nations. We require at least 500 respondents for each question within each country
and wave. These criteria yield 408 longitudinal variables across 72
countries. Appendix~\ref{app:data_sources} lists the topic categories.

\subsection{Gold Response Distribution}
For each \emph{variable} in a given country and year, we remove invalid responses such as ``don't know'' and calculate the proportion of people
choosing each response option. The survey sample may differ in
demographic composition from the country's population. To account
for this, we use survey-provided weights to adjust the response
proportions so that overrepresented or underrepresented demographic
groups better match their shares in the national population.
These weighted proportions form the gold response distribution $G_t$, that represents the population's responses for that question, country, and survey year $t$. Appendix~\ref{app:gold_distributions} provides details
on the survey weights and the calculation.

\subsection{Temporal Pairs}
\label{subsec:temporal_pairs}
For each variable observed in a country at both an older and a newer time point, we create one \emph{temporal pair}. Each pair consists of two instances of the same survey question for
the same country: one specifies the older survey year and the other
specifies the newer survey year. We denote the corresponding gold response distributions by $G_{\mathrm{old}}$ and $G_{\mathrm{new}}$, respectively. The newer year is the most recent year (ISSP) or Wave 7 (WVS), and the older year is the oldest available year for that variable and country.
This yields 10,360 temporal pairs across 72 countries, thus creating 20,720 dated prompts overall. Figure~\ref{fig:temporal_pair_example} shows an example from ISSP.

\begin{figure}[!ht]
\centering
\footnotesize
\begin{tcolorbox}[width=\columnwidth, colback=gray!3, colframe=black!55,
  boxrule=0.35pt, arc=1pt, left=4pt, right=4pt, top=4pt, bottom=4pt,
  segmentation style={black!30}]
\textbf{Example temporal pair (France)}

\vspace{0.4em}
\begin{tabular}{@{}p{0.48\linewidth}p{0.48\linewidth}@{}}
\textbf{Older prompt} & \textbf{Newer prompt} \\
In France, during \textbf{2010}, \dots & In France, during \textbf{2020}, \dots \\
\end{tabular}

\vspace{0.4em}
\emph{Only the year changes: the question, response options, and
instructions are identical.}

\tcbline

\textbf{Question:} How much do you agree or disagree: ``We worry too much
about the future of the environment and not enough about prices and
jobs today.''

\vspace{0.3em}
\textbf{Gold response distributions (\%)}
\begin{center}
\setlength{\tabcolsep}{6pt}
\begin{tabular}{lrr}
\toprule
\textbf{Response option} & \textbf{Older (2010)} & \textbf{Newer (2020)} \\
\midrule
Strongly agree    & 16.5 &  3.0 \\
Agree             & 30.0 & 14.4 \\
Neither           & 21.6 & 21.2 \\
Disagree          & 21.7 & 31.8 \\
Strongly disagree & 10.2 & 29.6 \\
\bottomrule
\end{tabular}
\end{center}
\end{tcolorbox}
\caption{Example temporal pair from ISSP: two prompts that differ only in
the year, each with its own gold response distribution.}
\label{fig:temporal_pair_example}
\end{figure}

\subsection{Categorization of Temporal Pairs}
\label{subsec:temporal-pair-categorize}

To evaluate the temporal awareness of LLMs, we first measure how much cultural attitudes changed within each temporal pair.
This allows us to assess whether models capture the extent
of that change. We measure this change using the Wasserstein
distance \citep{villani2009optimal,
peyre2019computational} between $G_{\mathrm{old}}$ and $G_{\mathrm{new}}$, which takes the order of
response options into account.
Because questions have different numbers of response options,
we normalize the distance to a common $[0,1]$ range.
We denote this normalized Wasserstein distance by $W$ and define the
\emph{human shift}, $d_h$, as:
\begin{equation}
d_h = W\left(G_{\mathrm{old}}, G_{\mathrm{new}}\right).
\label{eq:human_shift}
\end{equation}
Larger values indicate greater changes in cultural attitudes.
Appendix~\ref{app:wasserstein_distance} provides the calculation.

We use $d_h$ to group temporal pairs into four categories: stable, moderate, strong, and very strong shifts. This allows us to compare model behavior across different levels of cultural change. Table~\ref{tab:shift_classification_counts} reports the thresholds and the number of pairs in each category, and
Appendix~\ref{sec:threshold_robustness} shows that our findings hold
under alternative thresholds.

\begin{table}[!ht]
\centering
\small
\begin{tabular}{llr}
\toprule
\textbf{Shift category} & \textbf{Human shift $d_h$} & \textbf{Total pairs} \\
\midrule
Stable & $d_h \leq 0.03$ & 2,607 \\
Moderate shift & $0.03 < d_h < 0.10$ & 5,535 \\
Strong shift & $0.10 \leq d_h < 0.20$ & 1,828 \\
Very strong shift & $d_h \geq 0.20$ & 390 \\
\midrule
Total & & 10,360 \\
\bottomrule
\end{tabular}
\caption{Human-shift thresholds and temporal-pair counts for each category.}
\label{tab:shift_classification_counts}
\end{table}

\begin{figure}[!ht]
\centering
\footnotesize
\begin{tcolorbox}[width=\columnwidth, colback=gray!3, colframe=black!55,
  boxrule=0.35pt, arc=1pt, left=4pt, right=4pt, top=3pt, bottom=3pt]
\raggedright
\setlength{\tabcolsep}{5pt}
\renewcommand{\arraystretch}{0.92}
\textbf{(a) False negative:} large shift, unchanged mode\\[3pt]
\emph{Bangladesh: ``Is having the army rule a good or bad way of
governing this country?''}
\begin{center}
\begin{tabular}[c]{@{}cl rr@{}}
\toprule
& & \textbf{1996} & \textbf{2018} \\
\midrule
1 & Very good   &  1.0 & 13.1 \\
2 & Fairly good &  5.8 & 25.9 \\
3 & Fairly bad  & 23.1 & 22.2 \\
4 & Very bad    & \textbf{70.1} & \textbf{38.9} \\
\bottomrule
\end{tabular}%
\hspace{1.2em}%
\begin{tabular}[c]{@{}c@{}}$W = 0.251$\end{tabular}
\end{center}
\vspace{0.2em}
\textbf{(b) False positive:} little shift, changed mode\\[3pt]
\emph{United States: ``How much conflict is there between the
unemployed and people with jobs?''}
\begin{center}
\begin{tabular}[c]{@{}cl rr@{}}
\toprule
& & \textbf{1987} & \textbf{1992} \\
\midrule
1 & Very strong conflict &  9.8 & 12.6 \\
2 & Strong conflict      & 39.2 & \textbf{40.4} \\
3 & Not very strong      & \textbf{43.3} & 38.8 \\
4 & No conflict          &  7.7 &  8.2 \\
\bottomrule
\end{tabular}%
\hspace{1.2em}%
\begin{tabular}[c]{@{}c@{}}$W = 0.024$\end{tabular}
\end{center}
\end{tcolorbox}
\caption{The most common answer (mode) can misrepresent cultural
change: (a) ``Very bad'' falls from 70.1\% to 38.9\% but remains the
mode. (b) mode changes despite little distributional change. Bold marks the mode.}
\label{fig:mcq_limits}
\end{figure}

\subsection{Why Distribution Evaluation?}
\label{subsec:distribution evaluation}
Following our second design principle, we evaluate cultural change using full response distributions, since evaluations based on a single correct answer may miss shifts in population attitudes. To examine this limitation, we construct an MCQ version of our benchmark, treating each response option as a choice and the most common population response (the mode) as the correct answer.

Figure~\ref{fig:mcq_limits} illustrates two limitations. A substantial shift in the response distribution may leave the mode unchanged, causing the MCQ evaluation to miss the shift (false negative). Conversely, a slight shift may change the mode, causing the evaluation to overstate the change (false positive). Thus, modal answers alone cannot reliably measure the magnitude of cultural change.

Both cases are frequent in our data. Among pairs whose distribution has
shifted, 68.2\% keep the same mode in both periods, and 8.5\% of stable
pairs change mode despite little distributional change. A benchmark
scored on modal answers would therefore misrepresent how culture changed
for most of our data. For this reason, we evaluate full response
distributions, which account for changes across all response options.
Appendix~\ref{app:mcq} reports model results for the MCQ format.
\subsection{Prompt Construction and Prediction}
\label{sec:prompt_construction}
Recent work has explored direct prompting of response distributions from LLMs, asking models to output numerical distributions over answer options in structured formats \citep{suzuki-etal-2025-reliability, meister-etal-2025-benchmarking}. \citet{meister-etal-2025-benchmarking} compare multiple methods of distribution extraction, including verbalized distributions, sampling-based outputs, and model log-probabilities. They show that directly prompting an LLM for a response distribution outperforms repeated individual querying \citep{meister-etal-2025-benchmarking, gong2026characterizing}. We adopt this direct distribution-prompting approach. 

For each temporal pair, we construct two prompts, one for the older year and one for the newer year. Each prompt includes the country, year, survey question, and response options. We ask models to predict the percentage of people who would choose each option. Figure~\ref{fig:temporal_pair_example} shows the question pair, Appendix~\ref{app:prompt_example} provides the full prompt.

We denote the prediction for survey year $t$ by $P_t$. An accurate prediction should closely match the gold distribution $G_t$, giving a small normalized Wasserstein distance $W(P_t, G_t)$. For each temporal pair, we denote the predictions for the older and newer years by $P_{\mathrm{old}}$ and $P_{\mathrm{new}}$, respectively.

\section{Experiments and Results}
\label{sec:exp}

A temporally aware model should reflect cultural attitudes as they change over time, while producing similar predictions when those attitudes remain stable. We first apply a knowledge filter to retain questions for which models show baseline cultural knowledge (Section~\ref{sec:4.1}). We then present three Research Findings (RFs): (1) predicting both periods accurately becomes harder as cultural change increases (Section~\ref{sec:rq1}), (2) models follow the direction of larger cultural shifts but underestimate their magnitude (Section~\ref{sec:rq2}) and (3) naming a year does not consistently improve alignment (Section~\ref{sec:rq3}). Finally, in Section~\ref{sec:topics}, we examine how temporal performance varies across topics and regions.

We evaluate six LLMs using the same prompts and applying the knowledge
filter separately to each: Llama-3.3-70B \citep{grattafiori2024llama},
GPT-4o \citep{hurst2024gpt4o}, GPT-5.4 \citep{openai2026gpt54},
Qwen3.5-35B-A3B \citep{qwen3.5}, Qwen3.8-Flash-Next
\citep{qwen3.8flashnext}, and DeepSeek-V3 \citep{deepseekai2024v3}.

\subsection{Cultural Knowledge Filter}
\label{sec:4.1}

Our goal is to evaluate whether models can accurately predict cultural attitudes across different time periods. However, poor predictions could also imply limited knowledge, rather than an inability to track how cultural attitudes change. Thus, before evaluating temporal awareness, we apply a \emph{filter} to only retain questions for which the model shows some evidence of baseline knowledge.

To apply this filter, we estimate the model's baseline knowledge by comparing its predictions with observed survey responses. For each question--country combination, we prompt the model to predict the response distribution without specifying a year, giving the \emph{undated prediction} $P_{\mathrm{undated}}$. We compute the Wasserstein distance between $P_{\mathrm{undated}}$ and the gold distribution for each available survey year, and define $K$ as the minimum of these distances. A small $K$ means the model's prediction aligns with the cultural attitudes of at least one year, suggesting some baseline knowledge.

We apply the knowledge filter separately to each model, retaining the 75\% of question--country combinations with the smallest $K$ values (approximately 7,700 temporal pairs per model). Appendix~\ref{app:gate} provides further details on the selection process and retention statistics.

\begin{figure*}[t!]
\centering
\includegraphics[width=0.9\textwidth]{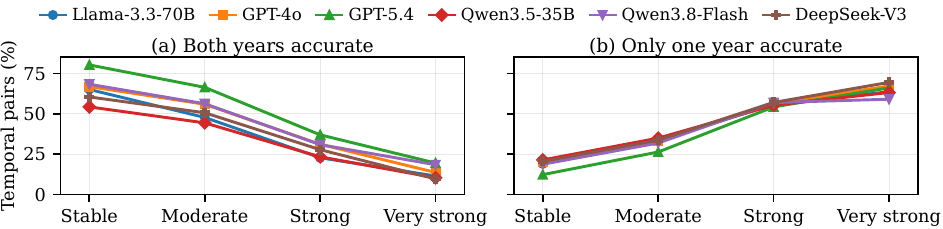}
\caption{As human shift increases, for all six models: (a) pairs with
both predictions within $\tau = 0.10$ of their gold distributions
decline (b) pairs with exactly one such prediction rise. Full results in Appendix~\ref{app:rq1}.}
\label{fig:rq1_trend}
\end{figure*}

\begin{figure}[t!]
\centering
\includegraphics[width=\columnwidth]{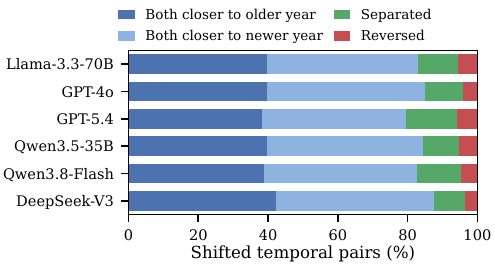}
\caption{In most shifted pairs (moderate or larger)%
, both of a model's predictions are closer to the same year's gold distribution (either shade of blue), rather than matching its own year (green). Reversed pairs (red) are rare. Breakdown by shift category in
Appendix~\ref{app:rq3_anchoring}.}
\label{fig:rq1_alignment}
\end{figure}

\subsection{RF1: Predicting Both Periods Accurately Becomes Harder as Cultural Change Increases} %
\label{sec:rq1}

We examine whether models accurately predict both periods of a
temporal pair. Across all six models, the share of pairs accurately
predicted in both periods declines as human shift increases
(Figure~\ref{fig:rq1_trend}), while the share with only one accurate
prediction rises. Moreover, both predictions are usually closer to
the same year's gold distribution than each to its own
(Figure~\ref{fig:rq1_alignment}).

\paragraph{Setup.} For each temporal pair, we obtain two predicted distributions: $P_{\mathrm{old}}$ for the older year and $P_{\mathrm{new}}$ for the newer year. We measure prediction errors using the distance $W$ to the corresponding gold distributions:
\begin{equation}
\begin{aligned}
d_{\mathrm{old}} &= W \left(P_{\mathrm{old}}, G_{\mathrm{old}}\right), \\
d_{\mathrm{new}} &= W \left(P_{\mathrm{new}}, G_{\mathrm{new}}\right).
\end{aligned}
\end{equation}

Averaging the prediction errors $d_{\mathrm{old}}$ and
$d_{\mathrm{new}}$ across temporal pairs does not clearly reveal whether
both predictions are accurate within the same pair. To assess
accuracy at the pair level, we compare these errors against
a tolerance of $\tau = 0.10$. Each pair is then classified
as \emph{both}, \emph{older only}, \emph{newer only}, or
\emph{neither}, depending on which predictions fall within
the tolerance. Appendix~\ref{app:rq1_tau} reports results
for other settings of $\tau$.

\paragraph{Results.}
Figure~\ref{fig:rq1_trend}(a) shows that, across all six models,
the proportion of temporal pairs where both $P_{\mathrm{old}}$
and $P_{\mathrm{new}}$ are accurate declines as human shift increases.
In contrast, the share of pairs where only one prediction is accurate rises,
exceeding 50\% for strong and very strong shifts
(Figure~\ref{fig:rq1_trend}(b)).

\paragraph{Alignment of predictions.} Accuracy alone does not show whether each prediction is closer to its own year's gold distribution than to the other's. To examine this, we compare each prediction with both $G_{\mathrm{old}}$ and $G_{\mathrm{new}}$. We focus on pairs with moderate or larger human shifts. Pairs are \textbf{separated} when each prediction is closer to its own year's gold distribution, and \textbf{reversed} when each is closer to the opposite year's. They are \textbf{older-aligned} or \textbf{newer-aligned} when both predictions are closer to $G_{\mathrm{old}}$ or $G_{\mathrm{new}}$, respectively.

\paragraph{Alignment results.}
Figure~\ref{fig:rq1_alignment} shows that, across all six models,
$P_{\mathrm{old}}$ and $P_{\mathrm{new}}$ are closer to the same
gold distribution for most shifted pairs. The \emph{older-aligned}
and \emph{newer-aligned} categories account for 80--87\% of these
pairs, compared with only 9--15\% in the \emph{separated} category
(Appendix~\ref{app:rq3_anchoring}). Although \emph{separated} pairs become more common as human shift increases, most pairs still align with the same gold distribution. \emph{Reversed} pairs are rare.

\begin{figure*}[t!]
\centering
\includegraphics[width=0.95\textwidth]{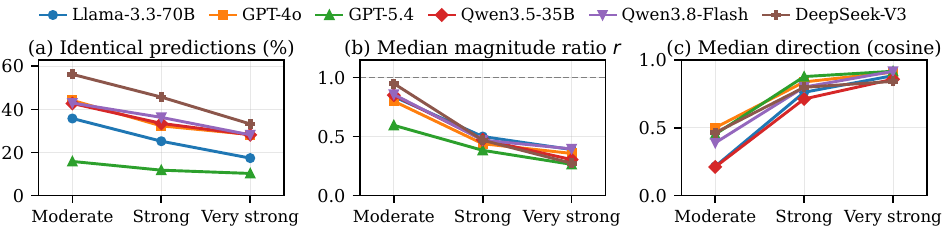}
\caption{As human shift increases, for all six models: (a) identical
predictions become less common but persist, (b) median magnitude ratio
$r=d_m/d_h$ decreases (dashed line: $r=1$): amount of observed change captured decreases (c) median cosine similarity: directional alignment improves. Full results in Table~\ref{tab:rq2}, Appendix~\ref{app:rq2_full}.}
\label{fig:rq2_trend}
\end{figure*}

\subsection{RF2: Models Follow the Direction of Larger Cultural Shifts but Underestimate Their Magnitude}
\label{sec:rq2}

RF1 showed that models are less likely to predict both periods accurately as human shift increases, while $P_{\mathrm{old}}$ and $P_{\mathrm{new}}$ often resemble the same gold distribution. We now examine whether the change between these predictions matches the observed cultural shift in magnitude and direction. The predicted change may be too small, absent, in the opposite
direction, or too large, Appendix~\ref{app:rq2_arrows} shows each case for Llama-3.3-70B.

Across all six models, $P_{\mathrm{old}}$ and $P_{\mathrm{new}}$ sometimes remain identical despite observed cultural change, although this becomes less common as human shift increases (Figure~\ref{fig:rq2_trend}(a)). When human shift increases, predictions capture a smaller proportion of the observed cultural change (Figure~\ref{fig:rq2_trend}(b)), while directional alignment improves (Figure~\ref{fig:rq2_trend}(c)). Thus, better directional alignment does not ensure that models capture the magnitude of cultural change.

\paragraph{Magnitude of change.}
For each temporal pair, we define the \emph{model shift}, $d_m$, as the normalized Wasserstein distance between $P_{\mathrm{old}}$ and $P_{\mathrm{new}}$:
\begin{equation}
d_m = W\left(P_{\mathrm{old}}, P_{\mathrm{new}}\right).
\label{eq:model_shift}
\end{equation}
To assess how much of the observed change the model captures, we compute the \emph{magnitude ratio} $r = d_m / d_h$, %
where $d_h$ is the human shift (Equation~\ref{eq:human_shift}). A ratio of $1$ indicates that the predicted and observed changes are equal in magnitude. Values below $1$ indicate underestimation, while values above $1$ indicate overestimation. We report median $r$ over all temporal pairs with non-zero model shift ($d_m > 0$).  Appendix~\ref{app:rq3_magnitude} reports median $r$ when pairs with identical model predictions are included.

\paragraph{Results for magnitude.}
Figure~\ref{fig:rq2_trend}(a) shows that identical predictions
($d_m = 0$) become less common as human shift increases, but
persist even for very strong shifts across all six models.
For pairs with non-zero model shift ($d_m > 0$), median
magnitude ratio $r$ remains below $1$, and decreases as human shift increases (Figure~\ref{fig:rq2_trend}(b)).
Even GPT-5.4, which produces identical predictions least often,
still underestimates observed cultural change.
Appendix~\ref{app:rq2_examples} shows examples of limited
model change, while Appendix~\ref{app:rq2_shift_plot}
compares $d_m$ and $d_h$ across pairs.
\begin{figure*}[t!]
\centering
\includegraphics[width=0.9\textwidth]{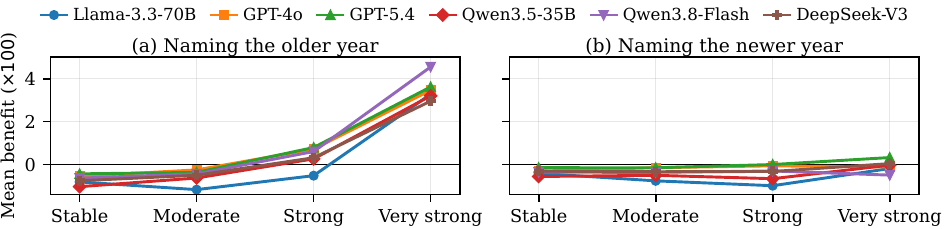}
\caption{Mean benefit (reduction in $W$, $\times 100$) by human shift. Positive values mean better alignment. (a) Naming the older year helps most for very strong shifts, (b) naming the newer year helps little. Full results in Appendix~\ref{app:rq2_outcomes}.}
\label{fig:rq3}
\end{figure*}

\paragraph{Direction of change.}
The magnitude ratio $r$ does not indicate whether the change from
$P_{\mathrm{old}}$ to $P_{\mathrm{new}}$ follows the same
direction as the change from $G_{\mathrm{old}}$ to $G_{\mathrm{new}}$. Because answer options are ordered (see Figure~\ref{fig:temporal_pair_example}), we use \emph{cumulative proportions} to describe changes along the response scale. At each option, we calculate the share of responses at or below that option. For example, on a scale ordered from \emph{strongly agree} to \emph{strongly disagree}, the cumulative proportion at \emph{agree} is the combined share of responses from \emph{strongly agree} and \emph{agree}.

Subtracting the older year's cumulative proportions from the newer year's gives a vector showing how the distribution changes along the response scale. We construct this vector separately for the gold distributions and model predictions:
\begin{equation}
\begin{aligned}
\mathbf{g} &= F(G_{\mathrm{new}})-F(G_{\mathrm{old}}),\\
\mathbf{p} &= F(P_{\mathrm{new}})-F(P_{\mathrm{old}}),
\end{aligned}
\end{equation}
where $F$ denotes the vector of cumulative response proportions. We measure \emph{directional alignment} using cosine similarity between $\mathbf{g}$ and $\mathbf{p}$: values near $1$, $0$, and $-1$ indicate similar directions, little alignment, and opposing directions, respectively. We exclude pairs with identical model predictions (10--56\% of shifted pairs; Figure~\ref{fig:rq2_trend}(a)), since their direction of change is undefined. 

\paragraph{Results for direction.}
Among pairs with $d_m > 0$, median directional alignment increases with human shift across all models (Figure~\ref{fig:rq2_trend}), unlike the magnitude ratio. For stable temporal pairs, models also predict noticeable change in 29.9--35.5\% of cases, despite little observed change (Appendix~\ref{app:stable_pairs}).

\subsection{RF3: Naming a Year Does Not Consistently Improve Cultural Alignment}
\label{sec:rq3}

We examine which survey year a prediction most closely resembles when no year is mentioned in the prompt, and whether naming the intended year improves alignment with that year's gold distribution. Across all models, undated predictions resemble an earlier year rather than the newest year's gold distribution in over half of cases. Naming a year offers little or no improvement on average. However, for very strong cultural shifts, specifying the older year improves alignment on average (Figure~\ref{fig:rq3}a).

\paragraph{Default period.}
For each retained question--country combination, we find the survey
year whose $G_t$ is closest to $P_{\mathrm{undated}}$
(Section~\ref{sec:4.1}) across all available years, including intermediate years not used in temporal pairs. We define the \emph{default survey year} as the year with the smallest distance $W$ between $P_{\mathrm{undated}}$ and $G_t$. Next, we record whether the \emph{default survey year} is the oldest, an intermediate, or the newest available survey year.

\paragraph{Results: Undated predictions often resemble earlier years.}
Across all models, the newest year is the closest match in only 46--48\% of cases (Appendix~\ref{app:rq3_full}). In more than half of cases, predictions are closer to an earlier survey year. Thus, undated predictions do not necessarily reflect the most recent cultural attitudes. 

\paragraph{Effect of naming a period.}
To evaluate the importance of temporal cues, we next ask whether specifying a year improves predictions for that period. Each temporal pair contains two dated prompts, one older and one newer, which we call \emph{items}. For each item, we compare the dated prediction $P_t$ and the undated prediction $P_{\mathrm{undated}}$ against the same gold distribution $G_t$, and define the \emph{benefit} of specifying year $t \in \{\text{old}, \text{new}\}$ as:
\begin{equation}
b_t = W(P_{\mathrm{undated}}, G_t) - W(P_t, G_t)
\end{equation}
A positive benefit means naming the year reduces prediction error, and a negative benefit means it increases error. We label each item \emph{helped}, \emph{hurt}, or \emph{unchanged} by the sign of its benefit, and report mean benefit separately for older and newer items.

\paragraph{Results: Limited overall benefit from specifying a year.}
Across models, specifying a year helps 20--44\% of items but hurts a similar share, so mean benefit is near zero or slightly negative. $P_t$
is also identical to $P_{\mathrm{undated}}$ for 11--54\% of items, more
often for newer items (Appendix~\ref{app:rq3_overall}).

We next examine mean benefit by human shift magnitude (Figure~\ref{fig:rq3}). For stable and moderate shifts, mean benefit is slightly negative for both older and newer items across all six models, so specifying a year offers no advantage. As human shift grows, however, specifying the older year begins to help: mean benefit for older items is positive in most models for strong shifts and in all six for very strong shifts. By contrast, specifying the newer year offers little benefit across shift levels.%

\subsection{Findings by Topic and Region}
\label{sec:topics}
We now highlight key findings on temporal awareness across topics and regions. Appendix~\ref{app:topics} provides the full results.

For strong and very strong cultural shifts, predictions are less accurate for political and economic topics than for family and personal life. For example, both-period accuracy for democracy and social inequality is roughly half that for family and life priorities. Likewise, models predict the observed direction of change in only about 50\% of cases for democracy and social inequality, compared with 82--94\% for family and life priorities. Notably, democracy is among the most accurately predicted WVS topics for stable pairs (71\%), yet its accuracy drops to 17\% for larger shifts.

Regional differences grow with the magnitude of cultural shift. For stable WVS pairs, we find that Western countries, the Middle East and North Africa (MENA), and Sub-Saharan Africa show similar accuracy (71--73\%). Under strong and very strong shifts, however, both-period accuracy is higher in Western countries (35\%) than in MENA and South and Southeast Asia (16\% on average). This gap between countries under larger shifts is smaller in ISSP, with 30\% accuracy in Western countries versus 23--29\% in other regions. Thus, models predict stable attitudes quite accurately across all regions, but for larger shifts, accuracy declines more sharply outside Western countries.

\section{Conclusion}
\label{sec:discussion}

We asked whether language models can represent not only what a
population believes, but also how those beliefs change over time.
Across six models, this ability remains limited. Models often capture a
country's attitudes at one point in time, yet present its evolving
culture as largely static: their predictions for different years tend
to stay close together. Furthermore, when attitudes shift strongly, models
capture the direction of change better than its magnitude. This gap
appears across all topics and regions, and is widest for political and
economic topics and for many non-Western populations.

The gap matters because LLMs are increasingly used to simulate survey
responses \citep{argyle2023out}. Naming a year, the most direct way to
ask about a period, does not reliably help. A model matching one
survey wave may appear culturally aware while missing how that
culture has changed. ChronoCulture makes this distinction measurable
by comparing full response distributions for the same question and
country across time.

Our findings point to several directions for improving temporal
cultural awareness. Future work could explore training on temporal
sources, learning to predict changes between survey response
distributions, and expanding data coverage for populations where models
show weaker temporal alignment. Together, these directions could help
models better represent cultures as they evolve
over time.

\section{Limitations}
ChronoCulture evaluates cultural attitudes using longitudinal survey data from the WVS and ISSP, so its coverage is limited to the countries, topics, and years included in these programs. At the time of writing, Wave 8 of the WVS has not been released, and our benchmark therefore does not capture changes in cultural attitudes after 2022. We also focus on questions with consistent, ordered response options across survey years, examining changes in reported attitudes and values rather than other aspects of culture, such as everyday customs and material culture.

We represent each population by its country-level response distribution. Although these distributions preserve the diversity of opinions within a country, they can hide differences between demographic, regional, or cultural groups within countries. Future extensions of ChronoCulture could incorporate
newer survey waves and finer-grained population data to examine
cultural change across more periods and communities.

\bibliography{custom}
\FloatBarrier

\appendix
\section{Benchmark}
\label{sec:appendix}
\subsection{Data Sources and Topic Coverage}
\label{app:data_sources}

The WVS contains seven 3--5 year waves of survey data from 1981 to 2022. We use waves 3--7, covering 1995 to 2022, because these waves provide repeated survey items across many countries and allow older-to-newer temporal comparisons. The years covered by each wave are listed in Table \ref{tab:wvs_waves}.

\begin{table}[!htbp]
\centering
\small
\begin{tabular}{ll}
\toprule
WVS wave & Fieldwork years \\
\midrule
Wave 3 & 1995--1998 \\
Wave 4 & 1999--2004 \\
Wave 5 & 2005--2009 \\
Wave 6 & 2010--2014 \\
Wave 7 & 2017--2022 \\
\bottomrule
\end{tabular}
\caption{World Values Survey waves used as temporal survey context.}
\label{tab:wvs_waves}
\end{table}
Each annual ISSP module focuses on one topic. Topics are repeated roughly every decade, and ISSP provides harmonized longitudinal datasets that make it possible to compare responses to the same questions across different time periods, covering 1985 to 2022 \citep{ZA5960,ZA8790,ZA8792,ZA8793,ZA8794,ZA8795,ZA8797,ZA4747}. The years covered by each topic are listed in Table \ref{tab:issp_topic_years}.
\begin{table}[!htbp]
\centering
\small
\begin{tabular}{ll}
\toprule
ISSP topic module & Survey years in benchmark \\
\midrule
Role of Government & 1985, 1990, 1996, 2006, 2016 \\
Social Inequality & 1987, 1992, 1999, 2009, 2019 \\
Family and Gender Roles & 1988, 1994, 2002, 2012, 2022 \\
Work Orientations & 1989, 1997, 2005, 2015 \\
Religion & 1991, 1998, 2008, 2018 \\
Environment & 1993, 2000, 2010, 2020 \\
National Identity & 1995, 2003, 2013 \\
Health and Health Care & 2011, 2021 \\
\bottomrule
\end{tabular}
\caption{ISSP repeated-topic modules represented in the benchmark.}
\label{tab:issp_topic_years}
\end{table}

Table \ref{tab:question_categories} details our questions by broad topic category, harmonized across WVS and ISSP since the two programs use different topic label wording.
\begin{table}[!htbp]
\centering
\small
\begin{tabular}{lr}
\toprule
Topic category & Questions \\
\midrule
Government, democracy, and institutions & 106 \\
Religion and national identity & 99 \\
Social inequality, work, and economic values & 83 \\
Family, health, and social attitudes & 74 \\
Environment, science, and technology & 46 \\
\midrule
Total & 408 \\
\bottomrule
\end{tabular}
\caption{Question distribution by broad topic category.}
\label{tab:question_categories}
\end{table}

\FloatBarrier

\subsection{Computing Gold Response Distributions}
\label{app:gold_distributions}

For each survey question, country, and year, we remove invalid
responses such as ``don't know.'' We then use the respondent
weights provided by the survey to calculate the proportion
choosing each response option:
\[
\text{Proportion}
=
\frac{
\begin{array}{c}
\text{Total weight of respondents}\\
\text{choosing the option}
\end{array}
}{
\begin{array}{c}
\text{Total weight of respondents}\\
\text{with valid answers}
\end{array}
}.
\]
These proportions sum to one and together form the gold
response distribution.

For illustration, consider three respondents with valid
answers to a question:

\begin{center}
\begin{tabular}{lll}
\hline
Respondent & Answer & Weight \\
\hline
Person 1 & Agree & 1 \\
Person 2 & Agree & 1 \\
Person 3 & Disagree & 2 \\
\hline
\end{tabular}
\end{center}

The total weight for ``Agree'' is $1+1=2$, and the total
weight for ``Disagree'' is $2$. Dividing each by the total
weight of $4$ gives:

\begin{center}
\begin{tabular}{lll}
\hline
Response option & Calculation & Proportion \\
\hline
Agree & $2/4$ & $50\%$ \\
Disagree & $2/4$ & $50\%$ \\
\hline
\end{tabular}
\end{center}

Although two respondents selected ``Agree'' and one selected
``Disagree,'' the third respondent's weight makes their answer
count twice as much as either of the others. The resulting
gold response distribution therefore assigns equal proportions
to both options.

\FloatBarrier
\subsection{Wasserstein Distance Metric}
\label{app:wasserstein_distance}
We calculate our normalized Wasserstein distance metric to compare distributions as follows:

For two distributions \(p\) and \(q\) over ordered response options
\(a_1 < a_2 < \cdots < a_K\), let the cumulative distribution be

\begin{equation}
F_p(a_j) = \sum_{\ell=1}^{j} p(a_\ell).
\label{eq:cdf}
\end{equation}

Using the above definition, the one-dimensional Wasserstein distance is

\begin{equation}
W_1(p,q)
=
\sum_{j=1}^{K-1}
\left|F_p(a_j)-F_q(a_j)\right|
(a_{j+1}-a_j).
\label{eq:wasserstein}
\end{equation}

Because survey questions use different scale lengths, we normalize by the maximum possible distance on the scale to calculate a normalized distance $W$:

\begin{equation}
W(p,q)
=
\frac{W_1(p,q)}{a_K-a_1}.
\label{eq:normalized_wasserstein}
\end{equation}
\FloatBarrier
\subsection{Prompt Example}
\label{app:prompt_example}
Figure \ref{fig:prompt_example} portrays an example of our verbalized distribution prompt style. Each temporal pair consists of two prompts. The newer-period prompt for this example is identical except that the target year is 2020, and the corresponding undated prompts detailed in Section~\ref{sec:4.1} simply exclude the year altogether. The same prompts were used for inference across all models.

\begin{figure}[!htbp]
\centering
\footnotesize
\begin{tcolorbox}[
    width=0.95\columnwidth,
    colback=gray!4,
    colframe=black!55,
    boxrule=0.4pt,
    arc=1pt,
    left=5pt,
    right=5pt,
    top=5pt,
    bottom=5pt
]
\textbf{Example prompt for distribution inference}

\vspace{0.4em}
In France in 2010, how much do you agree or disagree: We worry too much
about the future of the environment and not enough about prices and jobs
today. Use this response scale: 1 means `strongly agree', 2 means
`agree', 3 means `neither agree nor disagree', 4 means `disagree', and
5 means `strongly disagree'.

\vspace{0.4em}
Instead of choosing one answer, estimate the distribution of responses
among people in this country and time period.\\
Return percentages for every valid response option: 1, 2, 3, 4, 5. The
percentages must be non-negative and should sum to 100.\\
Return only a JSON object in exactly this key format:

\vspace{0.2em}
\texttt{\{"1": <percentage>, "2": <percentage>, "3": <percentage>,
"4": <percentage>, "5": <percentage>\}}
\end{tcolorbox}
\caption{Example prompt for distribution inference. The newer-year
prompt is identical except that the year is 2020.}
\label{fig:prompt_example}
\end{figure}

\FloatBarrier
\subsection{Modal-Answer Evaluation}
\label{app:mcq}

Section~\ref{subsec:distribution evaluation} argues that modal answers
cannot capture how survey distributions change. For completeness, we
also evaluate Llama-3.3-70B, our reference model, in a modal-answer
format. This analysis demonstrates its limitations.

\paragraph{Construction.}
We reuse the temporal pairs from Section~\ref{subsec:temporal_pairs},
but ask the model to choose a single answer instead of predicting a
distribution. The response options serve as answer choices, and the
mode of each gold distribution serves as the label
(Figure~\ref{fig:mcq_prompt}). A modal label is only meaningful when one
option is clearly ahead, so we keep only pairs in which the two most
common responses differ by at least 10 percentage points. We use stable
pairs and strong or very strong pairs whose modes change between
periods, and exclude moderate pairs, whose changes are usually spread
across the distribution without a change in mode. Analogous to the
knowledge filter, we then keep only pairs where the model answers at
least one period correctly. These filters leave 1,351 pairs (1,129
stable, 151 strong, and 71 very strong).

\paragraph{Results.}
Table~\ref{tab:mcq-accuracy} shows the share of pairs where the model 
selects the correct mode for both periods. Accuracy is high for stable
pairs (86.3\%) and falls sharply for shifted pairs (11.9\% for strong
and 14.8\% for very strong shifts), matching the distributional results
in Section~\ref{sec:rq1}. However, a correct label can hide substantial
error. In 24.4\% of shifted pairs, the model selects the correct mode
while capturing less than half of the distributional shift. Together
with the failure cases in Figure~\ref{fig:mcq_limits}, this is why we
treat modal-answer accuracy as a supplementary diagnostic rather than a
primary measure.

\begin{table}[!htbp]
\centering
\small
\setlength{\tabcolsep}{5pt}
\renewcommand{\arraystretch}{1.08}
\begin{tabular}{lr}
\toprule
\textbf{Gold survey shift} & \textbf{Both correct (\%)} \\
\midrule
Stable      & 86.3 \\
Strong      & 11.9 \\
Very strong & 14.8 \\
\bottomrule
\end{tabular}
\caption{Modal-answer accuracy (Llama-3.3-70B): accuracy falls sharply for shifted pairs. Both correct is the share of pairs where the model selects the correct mode for both periods.}
\label{tab:mcq-accuracy}
\end{table}

\begin{figure}[!htbp]
\centering
\footnotesize
\begin{tcolorbox}[
    width=0.95\columnwidth,
    colback=gray!4, colframe=black!55,
    boxrule=0.4pt, arc=1pt,
    left=5pt, right=5pt, top=5pt, bottom=5pt
]
\textbf{Example modal-answer prompt}

\vspace{0.4em}
In France, during \{time period\}, how much do most people agree or
disagree:

\emph{``We worry too much about the future of the environment and not
enough about prices and jobs today.''}

\vspace{0.3em}
\textbf{Choices.} (A) Strongly agree; (B) Agree; (C) Neither;
(D) Disagree; (E) Strongly disagree.

\vspace{0.3em}
\textbf{Task.} Choose the response option expected to be most common.
Return only one valid option code.

\vspace{0.3em}
\textbf{Gold answer.} 2010: B \quad 2020: D
\end{tcolorbox}
\caption{Example modal-answer prompt. The label is the modal response of
the corresponding survey distribution.}
\label{fig:mcq_prompt}
\end{figure}
\FloatBarrier

\section{Experimental Setup}

\subsection{Models and Inference Settings}
\label{app:models}
\paragraph{Inference configuration.} GPT-4o and GPT-5.4 were accessed through the OpenAI API. The other models were hosted locally with vLLM on NVIDIA H100 NVL GPUs: Llama-3.3-70B (vLLM 0.9.2, 2 GPUs), Qwen3.5-35B-A3B-FP8 and Qwen3.8-Flash-Next-FP8 (vLLM 0.30.0, 2 GPUs each), and DeepSeek-V3-0324-AWQ (vLLM 0.9.2, 4 GPUs). All models used temperature 0, top-$p$ 1.0, at most 120 output tokens, and one generation per prompt.

\paragraph{System prompt.} All models received the same system message: ``You estimate survey response distributions. Return only valid JSON. Do not include explanations, markdown, or extra text.''

\paragraph{Output parsing and normalization.} 
We instructed models to return a JSON object containing
response percentages for each provided answer option. If JSON parsing failed, we extracted percentage outputs using
regular expressions.

Parsed values were converted to floating-point numbers,
with negative values clipped to zero. We then normalized
the resulting values to sum to one, including when the
original percentages did not sum to 100. Outputs missing
response options, containing nonnumeric
values, or having a nonpositive or nonfinite total after
clipping were considered invalid. Appendix~\ref{app:prompt_robustness} shows, for Llama-3.3-70B, that repeated runs give nearly identical predictions and that our findings hold under alternative prompt instructions.

\subsection{Cultural Knowledge Filter: Scores and Retention}
\label{app:gate}

This section gives details of the filter described in
Section~\ref{sec:4.1}.

\paragraph{Nearest-survey distance.}
For each question--country combination, let $T$ be the set of all
available survey years, $G_t$ the gold distribution for year $t$, and
$P_{\mathrm{undated}}$ the model's prediction obtained without
specifying a year. We define
\begin{equation}
K = \min_{t \in T} W\left(P_{\mathrm{undated}}, G_t\right),
\label{eq:gate}
\end{equation}
the distance to the survey year that most closely matches the undated
prediction. The comparison includes every available survey year, not
only the two years of the temporal pair. A small $K$ means the undated
prediction closely matches at least one observed survey distribution.

\paragraph{Selection.}
Combinations with more available years have more chances to find a
close match, so $K$ tends to fall as coverage grows: for Llama-3.3-70B,
the median $K$ is $0.079$ for combinations with two available years and
$0.053$ for those with five. We therefore rank combinations separately
within each survey and within each group sharing the same number of
available years, and retain the 75\% with the smallest $K$ in each
group. The retained combinations from both surveys are then pooled.
Selection uses only undated predictions, and dated predictions are reserved
for the temporal evaluation.

\begin{figure}[!htbp]
    \centering
    \includegraphics[width=\columnwidth]{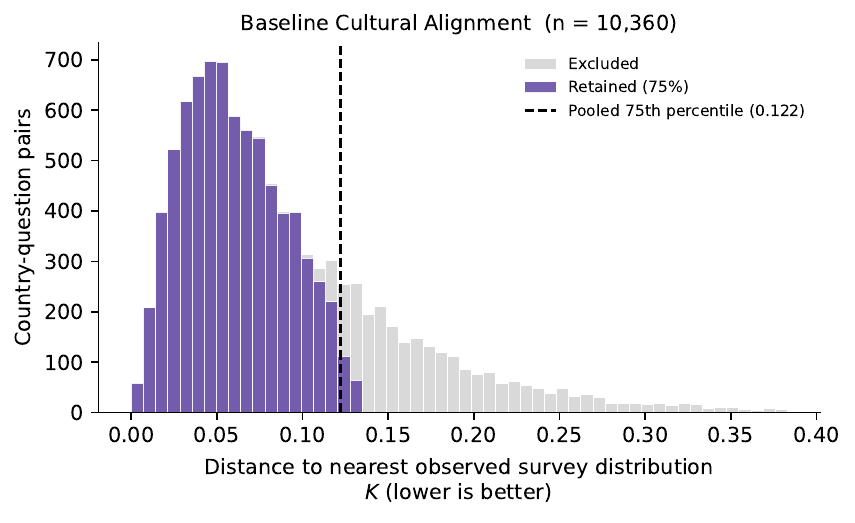}
    \caption{Distribution of the nearest-survey distance $K$ for
    Llama-3.3-70B across WVS and ISSP. Purple marks retained and gray
    excluded combinations. The dashed line marks the pooled 75th
    percentile ($0.122$) for reference only, since selection is done
    within survey and available-year groups.}
    \label{fig:gate_hist}
\end{figure}

\paragraph{Distribution and retention.}
Figure~\ref{fig:gate_hist} shows the distribution of $K$ for
Llama-3.3-70B. Most combinations fall below $0.12$, with a long tail
beyond $0.30$, where no survey year resembles the undated prediction.
The distribution has no natural break, so we choose a retention level. Table~\ref{tab:gate_retention} reports retention for all
models, and Table~\ref{tab:gate_sweep} shows how the shift categories
change at other retention levels. Below 50\%, the very strong category
falls under 100 pairs, too few for category-level comparison.

\begin{table}[!htbp]
\centering
\footnotesize
\setlength{\tabcolsep}{3pt}
\renewcommand{\arraystretch}{1.05}
\begin{tabular}{@{}lrrr@{}}
\toprule
\textbf{Model} & \textbf{WVS} & \textbf{ISSP}
& \textbf{Total} \\
\midrule
Llama-3.3-70B & 2,420/3,228 & 5,347/7,132 & 7,767 \\
GPT-4o        & 2,420/3,228 & 5,347/7,132 & 7,767 \\
GPT-5.4       & 2,420/3,228 & 5,347/7,132 & 7,767 \\
Qwen3.5-35B   & 2,420/3,228 & 5,347/7,132 & 7,767 \\
Qwen3.8-Flash & 2,420/3,228 & 5,347/7,132 & 7,767 \\
DeepSeek-V3   & 2,420/3,228 & 5,347/7,132 & 7,767 \\
\bottomrule
\end{tabular}
\caption{Knowledge-filter retention by model.
WVS and ISSP entries show retained/scored pairs, and Total shows pooled retained pairs.
Approximately 75\% are retained per survey.}
\label{tab:gate_retention}
\end{table}

\begin{table}[!htbp]
    \centering
    \small
    \setlength{\tabcolsep}{5pt}
    \begin{tabular}{lrrrrr}
        \toprule
        Retained & Stable & Mod. & Strong & V.\ Str. & Total \\
        \midrule
        100\% & 2,607 & 5,535 & 1,828 & 390 & 10,360 \\
        75\%  & 1,887 & 4,115 & 1,443 & 322 & 7,767 \\
        50\%  & 1,192 & 2,771 &   1,041 & 173 & 5,177 \\
        25\%  &   609 & 1,452 &   462 &  64 & 2,587 \\
        \bottomrule
    \end{tabular}
    \caption{Temporal pairs retained at each level for Llama-3.3-70B,
    by shift category.}
    \label{tab:gate_sweep}
\end{table}
\FloatBarrier
\section{RF1: Accuracy and Temporal Alignment}
\subsection{Full Period-Accuracy Results}
\label{app:rq1}
Table~\ref{tab:rq1} reports the full results summarized in
Figure~\ref{fig:rq1_trend}, including the split between older-only and
newer-only accuracy for each model. For reference, on the same retained pairs, a uniform prediction (equal
shares for every response option) is accurate for both periods in
8--11\% of stable pairs and in none of the very strong pairs. As a static reference, using each model's undated prediction for both years is accurate for both periods in 64--88\% of stable and 48--68\% of moderate pairs, above the dated predictions, but in only 16--25\% of strong and no very strong pairs, where dated predictions reach 23--37\% and 10--20\%. Bootstrap 95\% intervals over temporal pairs (1,000 resamples) extend about 2 points on either side of the estimate for stable and moderate pairs and about 4 points for very strong pairs, so the decline with human shift is much larger than sampling
variation.

\begin{table*}[!t]
\centering
\small
\setlength{\tabcolsep}{4pt}
\begin{tabular}{@{}lrrrrrrrrrrrr@{}}
\toprule
& \multicolumn{4}{c}{\textbf{Both}}
& \multicolumn{4}{c}{\textbf{Older only}}
& \multicolumn{4}{c}{\textbf{Newer only}} \\
\cmidrule(lr){2-5}\cmidrule(lr){6-9}\cmidrule(lr){10-13}
\textbf{Model} & Stab. & Mod. & Str. & V.\,str.
& Stab. & Mod. & Str. & V.\,str.
& Stab. & Mod. & Str. & V.\,str. \\
\midrule
Llama-3.3-70B  & 65.1 & 47.8 & 22.8 & 11.5 & 8.4 & 14.0 & 25.6 & 31.1 & 10.6 & 19.1 & 30.4 & 32.6 \\
GPT-4o         & 66.9 & 55.9 & 31.1 & 13.7 & 9.3 & 14.1 & 22.5 & 26.4 & 10.9 & 19.3 & 34.0 & 40.8 \\
GPT-5.4        & 80.3 & 66.4 & 37.0 & 19.5 & 4.8 & 10.5 & 23.7 & 27.7 &  7.6 & 15.9 & 30.6 & 38.4 \\
Qwen3.5-35B    & 54.3 & 44.4 & 23.4 & 10.4 & 9.6 & 16.2 & 24.3 & 27.3 & 11.9 & 18.8 & 31.0 & 35.9 \\
Qwen3.8-Flash  & 68.3 & 56.2 & 30.9 & 18.5 & 8.1 & 13.9 & 25.6 & 26.2 & 10.7 & 18.0 & 31.3 & 32.9 \\
DeepSeek-V3    & 60.5 & 50.7 & 27.8 &  9.7 & 7.9 & 15.1 & 25.2 & 29.9 & 12.6 & 18.6 & 31.9 & 39.6 \\
\bottomrule
\end{tabular}
\caption{RF1 accuracy by human shift: accuracy on both periods falls for every model, and newer only exceeds older only. Cells are the \% of temporal pairs whose predictions are within $\tau=0.10$ of gold for both, only the older, or only the newer period (Stab.\ = stable, Mod.\ = moderate, Str.\ = strong, V.\,str.\ = very strong).}
\label{tab:rq1}
\end{table*}

\subsection{Sensitivity to Accuracy Tolerance}
\label{app:rq1_tau}

\begin{figure*}[!t]
\centering
\includegraphics[width=0.9\textwidth]{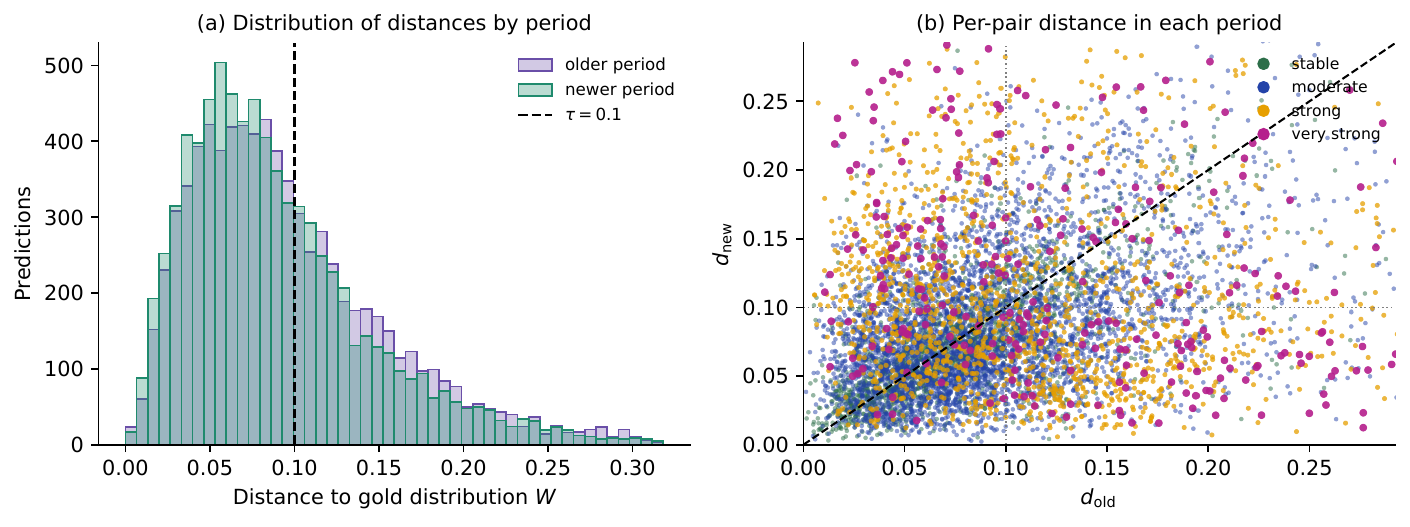}
\caption{Per-period prediction errors (Llama-3.3-70B): overall error distributions are similar, but many pairs have very different errors for the two periods. (a) Distances to gold, with $\tau=0.10$ marked. (b) $d_{\mathrm{old}}$ vs.\ $d_{\mathrm{new}}$ for each temporal pair; points far from the diagonal have a larger error for one period.}
\label{fig:rq1_figure}
\end{figure*}

We repeat the accuracy analysis from Section~\ref{sec:rq1} using four error tolerances: $\tau \in \{0.05, 0.10, 0.15, 0.20\}$. A prediction counts as accurate when its distance to the corresponding gold distribution is within $\tau$. A smaller tolerance therefore requires a closer match, while a larger tolerance allows a greater difference. For all six models, both-period accuracy falls from stable to very strong shifts at every tolerance. We report detailed results for Llama-3.3-70B. Figure~\ref{fig:rq1_figure} shows the scale of these distances: (a) the distribution of distances between each prediction and its gold distribution, with $\tau = 0.10$ marked, and (b) the distances for the older and newer predictions of each pair.

Table~\ref{tab:rq1_tau} reports accuracy for the older
and newer periods separately, along with the four
outcomes for each temporal pair. \emph{Both} means
that both predictions are accurate, \emph{older only}
and \emph{newer only} mean that only the prediction
for that period is accurate, and \emph{neither} means
that neither prediction is accurate.

\begin{table*}[!t]
\centering
\small
\setlength{\tabcolsep}{5pt}
\renewcommand{\arraystretch}{1.1}
\begin{tabular}{@{}clrrrrrr@{}}
\toprule
& & \multicolumn{2}{c}{\textbf{Accurate (\%)}}
& \multicolumn{4}{c}{\textbf{Pair outcomes (\%)}} \\
\cmidrule(lr){3-4}
\cmidrule(lr){5-8}
$\boldsymbol{\tau}$
& \textbf{Human shift}
& \textbf{Older}
& \textbf{Newer}
& \textbf{Both}
& \textbf{Older only}
& \textbf{Newer only}
& \textbf{Neither} \\
\midrule
0.05 & Stable      & 28.3 & 30.7 & 19.0 &  9.3 & 11.7 & 60.0 \\
     & Moderate    & 21.9 & 24.2 &  7.8 & 14.1 & 16.4 & 61.7 \\
     & Strong      & 18.2 & 20.0 &  2.6 & 15.6 & 17.4 & 64.4 \\
     & Very strong & 17.4 & 14.0 &  1.2 & 16.1 & 12.7 & 69.9 \\
     & All         & 22.6 & 24.6 &  9.3 & 13.3 & 15.3 & 62.1 \\
\midrule
0.10 & Stable      & 73.5 & 75.7 & 65.1 &  8.4 & 10.6 & 15.9 \\
     & Moderate    & 61.8 & 66.9 & 47.8 & 14.0 & 19.1 & 19.1 \\
     & Strong      & 48.4 & 53.2 & 22.8 & 25.6 & 30.4 & 21.3 \\
     & Very strong & 42.5 & 44.1 & 11.5 & 31.1 & 32.6 & 24.8 \\
     & All         & 61.3 & 65.5 & 45.8 & 15.5 & 19.7 & 19.0 \\
\midrule
0.15 & Stable      & 92.0 & 93.7 & 89.1 &  2.9 &  4.7 &  3.3 \\
     & Moderate    & 84.7 & 88.2 & 77.9 &  6.8 & 10.3 &  5.0 \\
     & Strong      & 72.3 & 76.4 & 55.6 & 16.7 & 20.8 &  6.9 \\
     & Very strong & 60.9 & 64.0 & 34.2 & 26.7 & 29.8 &  9.3 \\
     & All         & 83.2 & 86.4 & 74.7 &  8.5 & 11.7 &  5.1 \\
\midrule
0.20 & Stable      & 97.4 & 98.3 & 96.5 &  0.9 &  1.8 &  0.8 \\
     & Moderate    & 95.1 & 96.3 & 92.4 &  2.7 &  3.9 &  0.9 \\
     & Strong      & 88.4 & 88.5 & 78.4 &  10.0 &  10.1 &  1.5 \\
     & Very strong & 76.1 & 78.3 & 57.5 & 18.6 & 20.8 &  3.1 \\
     & All         & 93.6 & 94.6 & 89.3 &  4.3 &  5.3 &  1.1 \\
\bottomrule
\end{tabular}
\caption{RF1 accuracy across error tolerances (Llama-3.3-70B): at every tolerance, accuracy on both periods falls as human shift grows. Older and newer accuracy include pairs counted under both. The four pair outcomes are mutually exclusive and sum to about 100\% per row.}
\label{tab:rq1_tau}
\end{table*}

\subsection{Temporal Alignment}
\label{app:rq3_anchoring}

Table~\ref{tab:rq3_anchoring} reports the alignment
categories defined in Section~\ref{sec:rq1} for all models.
The analysis includes all retained moderate,
strong, and very strong human shift pairs,
including those with zero model shift.
Alignment depends on which gold distribution
each prediction more closely resembles, rather
than the accuracy tolerance used in
Section~\ref{sec:rq1}. Equal-distance cases are
identified using a numerical tolerance of
$10^{-12}$, none occurred.

Both predictions favor the same survey period
in most pairs. Correct separation becomes more
common as human shift increases, but remains
a minority outcome in every category. Reversed
pairs are comparatively uncommon throughout.

\begin{table*}[!t]
\centering
\small
\setlength{\tabcolsep}{4pt}
\begin{tabular}{@{}lrrrrrrrrrr@{}}
\toprule
& \multicolumn{3}{c}{\textbf{Separated (\%)}}
& \multicolumn{3}{c}{\textbf{Same period (\%)}}
& \multicolumn{4}{c}{\textbf{All shifted (\%)}} \\
\cmidrule(lr){2-4}\cmidrule(lr){5-7}\cmidrule(lr){8-11}
\textbf{Model} & Mod. & Str. & V.\,str. & Mod. & Str. & V.\,str.
& Sep. & Older-al. & Newer-al. & Rev. \\
\midrule
Llama-3.3-70B  &  8.8 & 15.2 & 26.7 & 85.5 & 79.1 & 68.9 & 11.3 & 39.8 & 43.2 & 5.7 \\
GPT-4o         &  8.7 & 15.0 & 23.9 & 87.2 & 81.0 & 73.9 & 11.1 & 39.8 & 45.1 & 4.0 \\
GPT-5.4        & 11.8 & 19.6 & 29.8 & 82.0 & 75.5 & 67.1 & 14.6 & 38.3 & 41.4 & 5.7 \\
Qwen3.5-35B    &  8.5 & 13.1 & 19.6 & 85.9 & 81.8 & 77.7 & 10.3 & 39.8 & 44.6 & 5.3 \\
Qwen3.8-Flash  &  10.0 & 16.0 & 27.9 & 84.9 & 79.7 & 69.5 & 12.4 & 38.9 & 43.9 & 4.8 \\
DeepSeek-V3    &  7.4 & 11.3 & 19.0 & 88.8 & 85.4 & 78.9 &  9.0 & 42.3 & 45.2 & 3.5 \\
\bottomrule
\end{tabular}
\caption{Period alignment of predictions: for every model, most shifted pairs align with a single period, and separated pairs remain a minority. Same period is older- plus newer-aligned; All shifted covers moderate to very strong pairs. Pairs with zero model shift are included.}
\label{tab:rq3_anchoring}
\end{table*}

\subsection{Results by Topic and Region}
\label{app:topics}

We break down all three analyses by survey topic and by world region to
test whether the main findings are confined to particular subjects or
populations. Topics follow each survey's own labels, so WVS and ISSP topics are reported separately.
Countries are grouped into seven regions: Western (Western Europe, North
America, Australia, and New Zealand), Eastern Europe and Central Asia,
Latin America, East Asia, South and Southeast Asia, the Middle East and
North Africa (MENA), and Sub-Saharan Africa. Because some groups contain
few very strong shifts, we combine strong and very strong shifts into a
single category, strong+. Each cell is computed per model on that model's
retained pairs, and cells with fewer than 20 pairs are left blank. We
report the mean across models and the range from the lowest to the
highest model. To measure whether models agree on which groups are
hardest, we rank the groups within each model and compute Kendall's $W$
across the six rankings, where $W=1$ means identical rankings. These breakdowns use the same pair-level results as Section~\ref{sec:exp}.

\begin{figure*}[!t]
  \centering
  \begin{minipage}{0.49\textwidth}
    \centering
    \includegraphics[width=\linewidth]{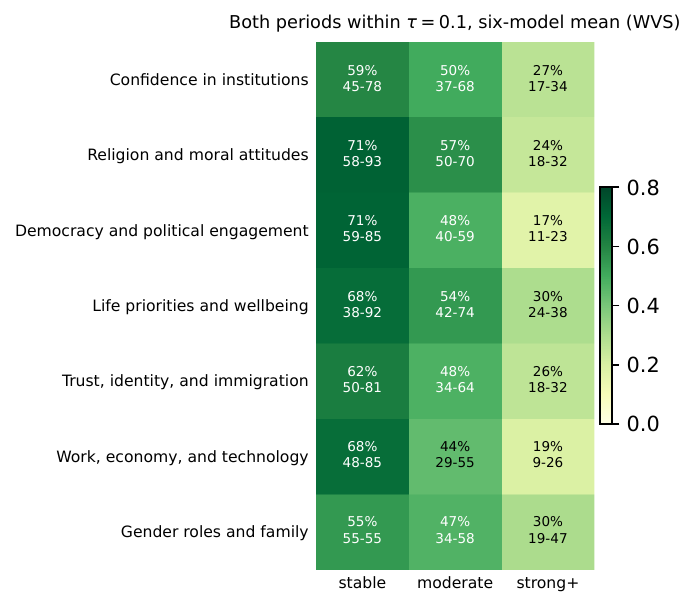}
  \end{minipage}\hfill
  \begin{minipage}{0.49\textwidth}
    \centering
    \includegraphics[width=\linewidth]{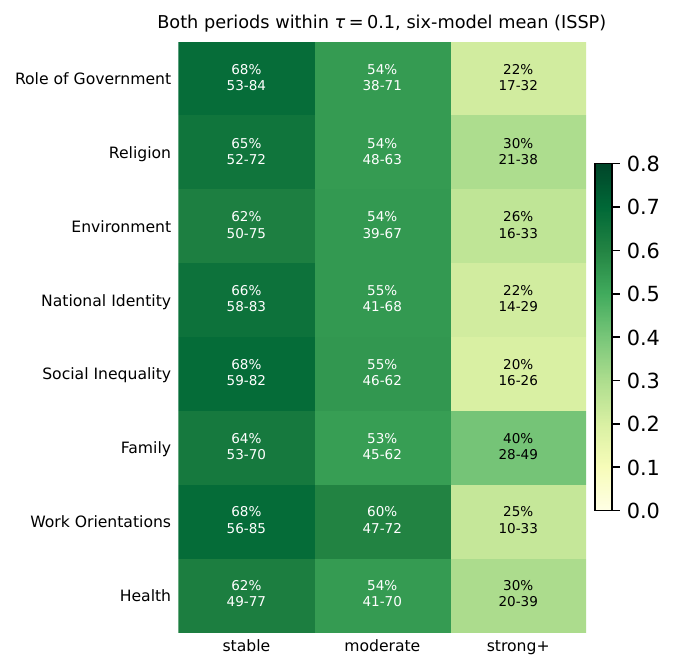}
  \end{minipage}
  \caption{Both-period accuracy by topic: political and economic topics are the hardest to track under large change, and family and personal life the easiest. Accuracy falls from stable to strong+ in every topic. Cells show the mean \% of pairs with both periods within $\tau=0.10$ across models with at least 20 pairs, with the model range below. Strong+ combines strong and very strong.}
  \label{fig:topic_heat}
\end{figure*}

\begin{figure*}[!t]
  \centering
  \begin{minipage}{0.49\textwidth}
    \centering
    \includegraphics[width=\linewidth]{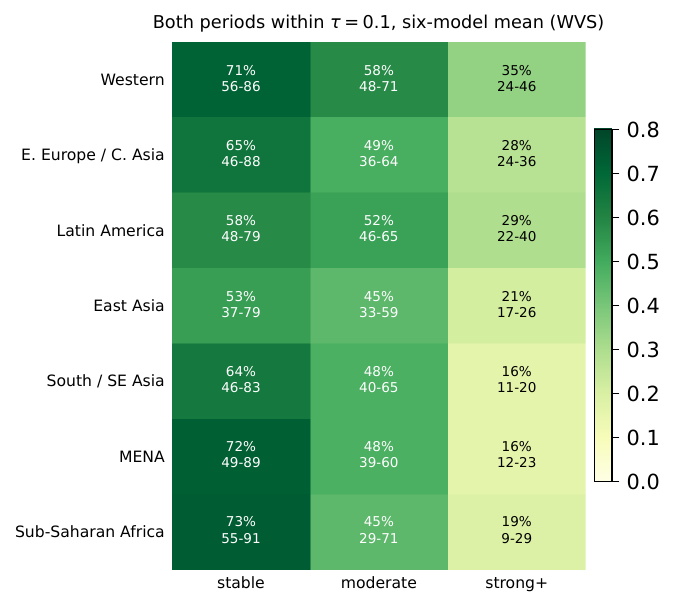}
  \end{minipage}\hfill
  \begin{minipage}{0.49\textwidth}
    \centering
    \includegraphics[width=\linewidth]{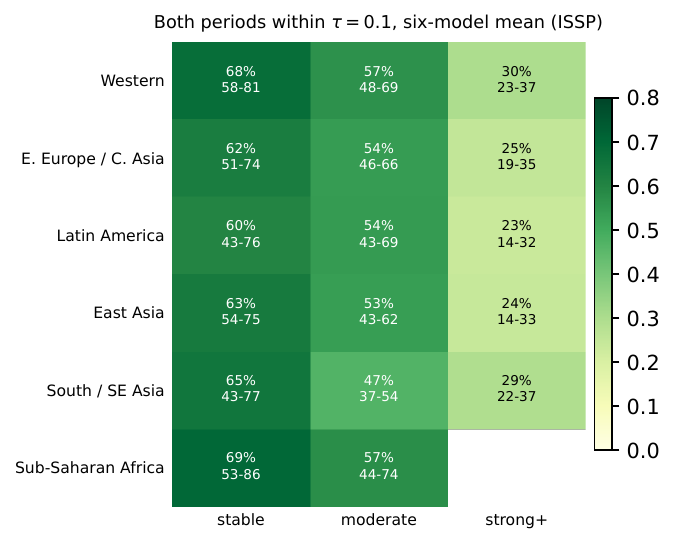}
  \end{minipage}
  \caption{Both-period accuracy by region: Western countries are tracked best under large change. In WVS, regions are similar for stable pairs, but for strong+ shifts Western countries are about twice as accurate as South and Southeast Asia and MENA. The gap is smaller in ISSP. Cells as in Figure~\ref{fig:topic_heat}.}
  \label{fig:region_heat}
\end{figure*}

\paragraph{Main findings hold in every group.}
For every model, topic, and region, both-period accuracy is lower for strong+ shift pairs than
for stable pairs
(Figures~\ref{fig:topic_heat} and~\ref{fig:region_heat}). In all but one
of the model, topic, and region combinations, the median $r$ is lower
for strong+ shifts than for moderate shifts, and for strong+ shifts
models capture between a third and two thirds of the observed change.
Naming a period changes accuracy little in every group. These patterns
are not explained by some groups changing more than others. The mean
human shift within strong+ is broadly similar across WVS topics
(0.15--0.17) and WVS regions (0.154--0.188), and ranges from 0.12 to 0.17
across ISSP topics.

\paragraph{Political and economic topics are the hardest.}
Under strong+ shifts, the least accurate topics concern politics and the
economy. In WVS, democracy and political engagement (17\%, range
11--23\%) and work, economy, and technology (19\%) rank last. In ISSP,
social inequality (20\%), role of government (22\%), and work
orientations (25\%) are among the least accurate. Topics about family and
personal life are more accurate: family (40\%) and health (30\%) in ISSP,
and life priorities and wellbeing (30\%) and gender roles and family
(30\%) in WVS. The same divide appears in RF2. On democracy and political
engagement, social inequality, and work orientations, models move in the
right direction for only 45--52\% of strong+ pairs, compared with
94\% for family and 82\% for life priorities. On
these political and economic topics, models also often leave their
prediction unchanged across periods, with zero model shift in 34--51\%
of strong+ pairs, compared with 11\% for family. Naming the older period
helps most for family ($+4.4$, $\times100$) and religion and moral
attitudes in WVS ($+2.6$), and slightly hurts for social inequality
($-1.0$).

These differences are not explained by the size of the shifts. For ISSP,
family is the most accurate topic even though its strong+ pairs have the
largest mean shift among ISSP topics (0.17). Health is the exception.
Its strong+ pairs have the smallest mean shift (0.12), which may partly
explain its accuracy. Topic difficulty also differs from topic
knowledge. Models disagree about which topics they represent best for
stable pairs ($W=0.16$ in WVS, $0.37$ in ISSP) but agree more closely on
which are hardest under strong+ shifts ($W=0.68$ and $0.75$). Democracy
and political engagement shows this most clearly. It is the second most
accurate WVS topic for stable pairs (71\%) and the least accurate for
strong+ shifts.

\paragraph{Western countries are tracked best under large change.}
In WVS, models are about equally accurate for Western countries (71\%),
MENA (72\%), and Sub-Saharan Africa (73\%) when attitudes are stable.
Under strong+ shifts, accuracy is 35\% for Western countries, 29\% for
Latin America, 28\% for Eastern Europe and Central Asia, 21\% for East
Asia, 19\% for Sub-Saharan Africa, and 16\% for both South and Southeast
Asia and MENA. Every model is more accurate for Western countries than
for South and Southeast Asia and MENA, and the regional ranking is
highly consistent across models ($W=0.83$). The gap remains when each
country is weighted equally (33\% for Western countries vs.\ 16--17\%
for South and Southeast Asia and MENA), so it is not driven by a few
countries with many pairs. The mean human shift within strong+ is
slightly larger for MENA (0.188) than for Western countries (0.154) but
similar for South and Southeast Asia (0.17). The same ordering appears
in RF2. Models move in the right direction for 75--77\% of strong+
pairs in Western countries and Latin America, compared with 58\% in
South and Southeast Asia and 43\% in Sub-Saharan Africa. Naming the
older period also helps most for Western countries ($+2.2$).

In ISSP, regional differences are smaller. Under strong+ shifts,
accuracy is 30\% for Western countries and 23--29\% for East Asia, Latin
America, Eastern Europe and Central Asia, and South and Southeast Asia,
and the regional ranking is less consistent across models ($W=0.32$).
Models still move in the right direction most often for Western
countries (82\% of strong+ pairs). Sub-Saharan Africa has too few
strong+ pairs for a model-level estimate, and MENA has too few pairs to
appear in the ISSP figure.

\paragraph{Summary.}
The difficulty of tracking change appears in every topic and region, but
it is not uniform. Models track change best for family and personal
life and for Western countries, and worst for political and economic
topics and for populations in South and Southeast Asia, MENA, and
Sub-Saharan Africa. Because models represent these populations about as
well as Western ones when attitudes are stable, the gap reflects how
well models track change rather than how well they know each population
at a single point in time.

\section{RF2: Magnitude, Direction, and Stability}
\subsection{Full Magnitude and Direction Results}
\label{app:rq2_full}

Table~\ref{tab:rq2} reports the full results summarized in
Figure~\ref{fig:rq2_trend}.

\begin{table*}[!t]
\centering
\small
\setlength{\tabcolsep}{5pt}
\begin{tabular}{@{}lrrrrrrrrr@{}}
\toprule
& \multicolumn{3}{c}{\textbf{Zero model shift (\%)}}
& \multicolumn{3}{c}{\textbf{Median ratio $r$}}
& \multicolumn{3}{c}{\textbf{Median cosine}} \\
\cmidrule(lr){2-4}\cmidrule(lr){5-7}\cmidrule(lr){8-10}
\textbf{Model} & Mod. & Str. & V.\,str. & Mod. & Str. & V.\,str. & Mod. & Str. & V.\,str. \\
\midrule
Llama-3.3-70B  & 35.8 & 25.2 & 17.4 & 0.84 & 0.50 & 0.39 & 0.22 & 0.76 & 0.88 \\
GPT-4o         & 44.2 & 32.3 & 28.3 & 0.80 & 0.44 & 0.36 & 0.50 & 0.84 & 0.91 \\
GPT-5.4        & 15.8 & 11.8 & 10.3 & 0.60 & 0.38 & 0.26 & 0.45 & 0.88 & 0.92 \\
Qwen3.5-35B    & 42.7 & 33.5 & 28.2 & 0.85 & 0.47 & 0.30 & 0.21 & 0.71 & 0.86 \\
Qwen3.8-Flash  & 43.0 & 36.3 & 28.2 & 0.86 & 0.47 & 0.40 & 0.39 & 0.80 & 0.91 \\
DeepSeek-V3    & 56.3 & 45.7 & 33.2 & 0.95 & 0.47 & 0.27 & 0.46 & 0.80 & 0.84 \\
\bottomrule
\end{tabular}
\caption{RF2 magnitude and direction by human shift: as shift grows, identical predictions become rarer, the median ratio falls, and the median cosine rises. Ratio and cosine use pairs with nonzero model shift.}
\label{tab:rq2}
\end{table*}

\subsection{Magnitude Ratios Including Identical Predictions}
\label{app:rq3_magnitude}

Section~\ref{sec:rq2} reports median magnitude ratios among pairs with nonzero model shift. Here, we also include pairs with identical predictions, which have a ratio of zero. Across all six models, the median ratio including these pairs is at most 0.46 in every human-shift category, so underestimation holds for every model. The decline across categories is clearest among pairs with nonzero model shift. Table~\ref{tab:rq3_magnitude} shows both medians for Llama-3.3-70B. Among pairs with nonzero model shift, its median ratio decreases as human shift increases, while including identical predictions it stays close to 0.30 in every category.

\begin{table}[!htbp]
\centering
\small
\setlength{\tabcolsep}{4pt}
\renewcommand{\arraystretch}{1.1}
\begin{tabular}{@{}lrr@{}}
\toprule
& \multicolumn{2}{c}{\textbf{Median ratio}} \\
\cmidrule(lr){2-3}
\textbf{Human shift}
& \shortstack{\textbf{Nonzero}\\\textbf{shift}}
& \textbf{All pairs} \\
\midrule
Moderate    & 0.838 & 0.320 \\
Strong      & 0.500 & 0.295 \\
Very strong & 0.390 & 0.304 \\
\bottomrule
\end{tabular}
\caption{Median magnitude ratio with and without identical predictions (Llama-3.3-70B): underestimation persists when identical predictions are included.}
\label{tab:rq3_magnitude}
\end{table}

\subsection{Stability Preservation and False Drift}
\label{app:stable_pairs}

The magnitude analysis in Section~\ref{sec:rq2} focuses on temporal
pairs whose cultural attitudes changed. Stable pairs require a different
comparison because their human shift is close to zero, making the
magnitude ratio less informative. We therefore examine the model shift
$d_m$ directly to determine whether the model also keeps its predictions
stable when the observed attitudes remain stable.

\textbf{Setup.}
We consider stable temporal pairs ($d_h \leq 0.03$). For each pair,
$d_m$, defined in Equation~\ref{eq:model_shift}, measures how much the
older and newer model predictions differ. We apply the same stability
boundary to the model shift. A pair preserves stability when
$d_m \leq 0.03$. When $d_m > 0.03$, the model predicts change beyond
this boundary even though the corresponding gold distributions remain
stable. We call this \emph{false drift}, distinguishing small false drift
($0.03 < d_m < 0.10$) from large false drift ($d_m \geq 0.10$). The 0.03 boundary matches the variation expected from survey sampling alone (Appendix~\ref{app:sampling_noise}). Appendix~\ref{app:rq4_by_survey} reports these results separately for WVS and ISSP.

\textbf{Results.}
Table~\ref{tab:false-drift-stable} shows that all six
models preserve stability in 64.5--70.1\% of stable
pairs. The remaining 29.9--35.5\% exhibit false drift,
with small false drift more common than large false
drift for every model. 
Thus, although stability preservation is the most common behavior, models still introduce noticeable change for a substantial fraction of
pairs whose observed attitudes remain stable.

These results complement the magnitude findings in
Section~\ref{sec:rq2}. There, models often underestimate change when
the gold distributions shift. Here, they can also introduce change when
the gold distributions remain stable. A small model shift on a stable
pair therefore should not by itself be interpreted as evidence that the
model has correctly recognized stability. More generally, the amount of
predicted change is not consistently calibrated to the amount of
observed cultural change.

\begin{table}[!htbp]
\centering
\small
\setlength{\tabcolsep}{4pt}
\begin{tabular}{@{}lrrr@{}}
\toprule
\textbf{Model} & \textbf{Preserves (\%)} & \textbf{Small (\%)} & \textbf{Large (\%)} \\
\midrule
Llama-3.3-70B  & 64.5 & 26.2 & 9.2 \\
GPT-4o         & 67.6 & 28.6 & 3.9 \\
GPT-5.4        & 67.0 & 30.2 & 2.8 \\
Qwen3.5-35B    & 68.6 & 21.9 & 9.5 \\
Qwen3.8-Flash  & 67.2 & 24.5 & 8.3 \\
DeepSeek-V3    & 70.1 & 22.9 & 7.0 \\
\bottomrule
\end{tabular}
\caption{Model behavior on stable pairs (\%): most stable pairs keep a steady prediction, but 29.9--35.5\% show false drift. Preserves: $d_m \leq 0.03$; small drift: $0.03 < d_m < 0.10$; large drift: $d_m \geq 0.10$.}
\label{tab:false-drift-stable}
\end{table}

\subsection{Stability Preservation by Survey}
\label{app:rq4_by_survey}

We separate the stable-pair analysis in
Appendix~\ref{app:stable_pairs} by survey to examine
whether the pooled results reflect similar
behavior in WVS and ISSP. Both surveys use
the same definitions: stable gold pairs have
$d_h \leq0.03$, and model predictions preserve
stability when $d_m\leq0.03$. Small false drift
corresponds to $0.03<d_m<0.10$, and large false
drift to $d_m\geq0.10$.

Table~\ref{tab:rq4_by_survey} shows that false
drift is more frequent in WVS. Stability
preservation exceeds combined false drift in
both surveys, but only narrowly in WVS
(50.6\% vs.\ 49.4\%). Large false drift
is also more than twice as frequent in WVS.
Since ISSP contributes more stable pairs,
the pooled percentages are closer to its
results.

\begin{table}[!htbp]
\centering
\small
\setlength{\tabcolsep}{4pt}
\renewcommand{\arraystretch}{1.1}
\begin{tabular}{@{}lrrr@{}}
\toprule
\textbf{Behavior} & \textbf{WVS} &
\textbf{ISSP} & \textbf{Pooled} \\
\midrule
Preserves stability & 50.6\% & 68.5\% & 64.5\% \\
Small false drift   & 33.9\% & 24.1\% & 26.2\% \\
Large false drift   & 15.5\% &  7.5\% &  9.2\% \\
\bottomrule
\end{tabular}
\caption{Stable-pair behavior by survey (Llama-3.3-70B): false drift is more frequent in WVS than in ISSP. Percentages may not sum to 100 due to rounding.}
\label{tab:rq4_by_survey}
\end{table}
\subsection{Human and Model Shift Across Temporal Pairs}
\label{app:rq2_shift_plot}

\begin{figure}[!ht]
    \centering
    \includegraphics[width=\columnwidth]
    {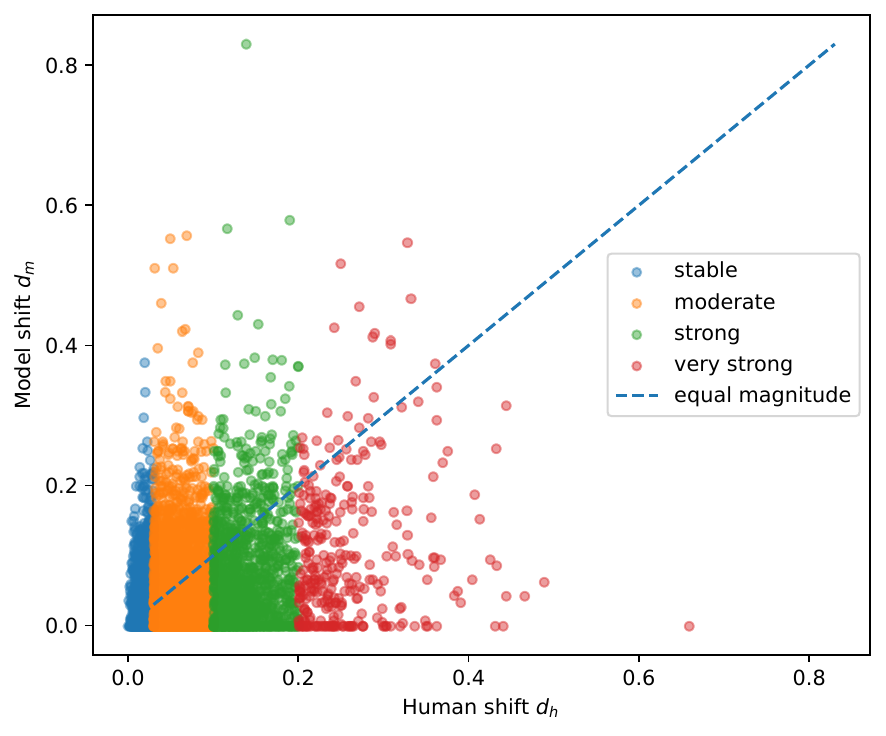}
    \caption{Human shift $d_h$ and model shift $d_m$
for Llama-3.3-70B on individual temporal pairs, colored by
    human shift category. The dashed diagonal
    marks equal magnitudes ($d_m=d_h$). Points below
    it indicate underestimation of the amount
    of cultural change.}
    \label{fig:rq2_human_model_shift}
\end{figure}

Figure~\ref{fig:rq2_human_model_shift} complements
the magnitude analysis in Section~\ref{sec:rq2}.
Each point represents a temporal pair, with human
shift $d_h$ on the horizontal axis and model shift
$d_m$ on the vertical axis. Points below the
diagonal indicate that the model predicts less
change than observed, while points above it
indicate more change. The diagonal represents
equal magnitudes but does not indicate whether
the direction of change is correct.
Stable pairs are included for context.

\begin{figure*}[!p]
    \centering
    \includegraphics[width=\textwidth]{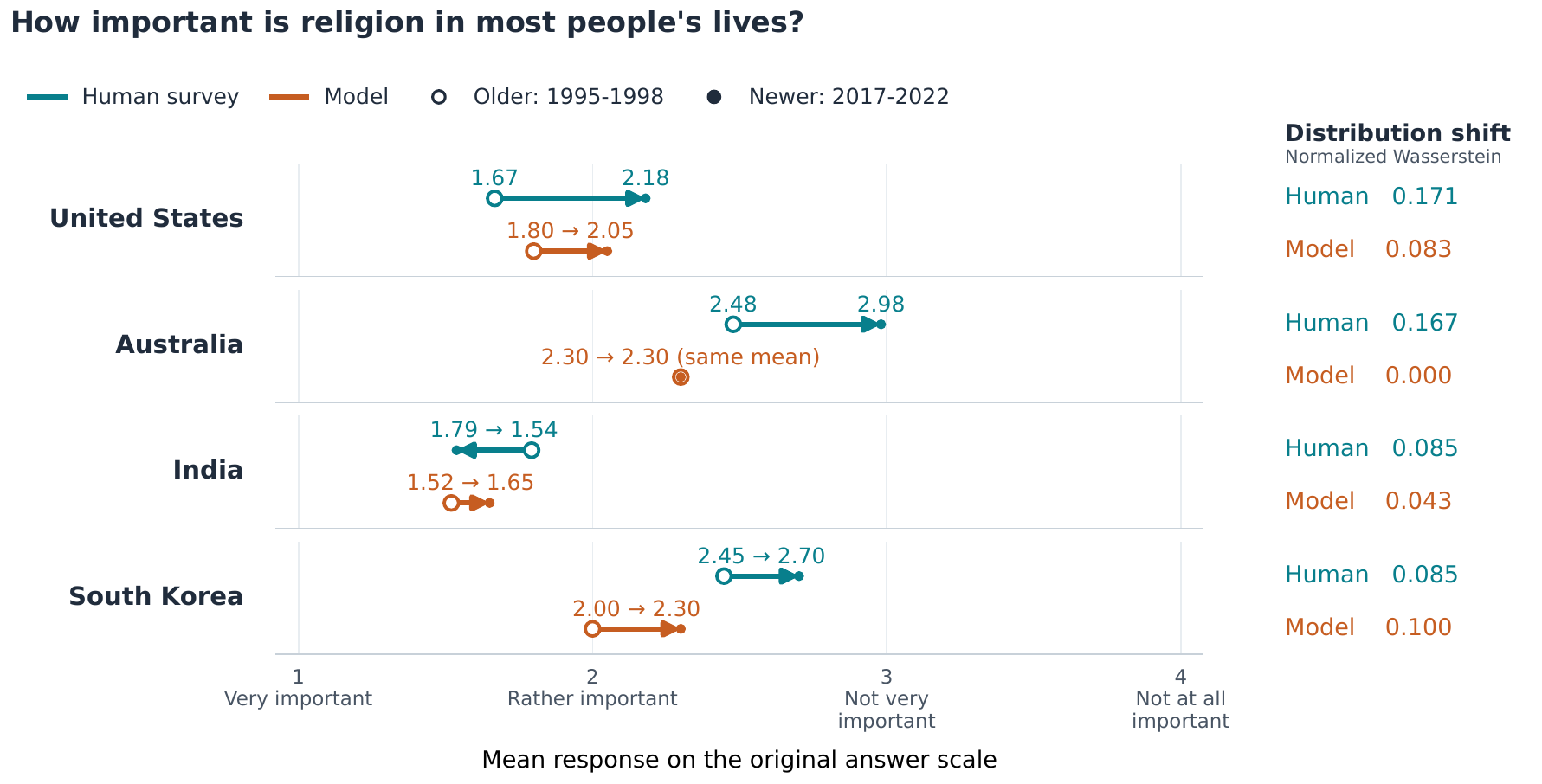}
    \caption{Four patterns of predicted change (Llama-3.3-70B). Arrows show
    the change in mean response from the older to the newer year. In the
    \textit{United States}, the model moves in the same direction as human
    responses but underestimates the change. In \textit{Australia}, it
    predicts no change despite an observed shift. In \textit{India}, model
    and human responses move in reverse direction, and in \textit{South
    Korea}, the model overestimates the change.}
\label{fig:rq2_arrows}
\end{figure*}

\begin{figure*}[!p]
    \centering
    \includegraphics[width=\textwidth]
    {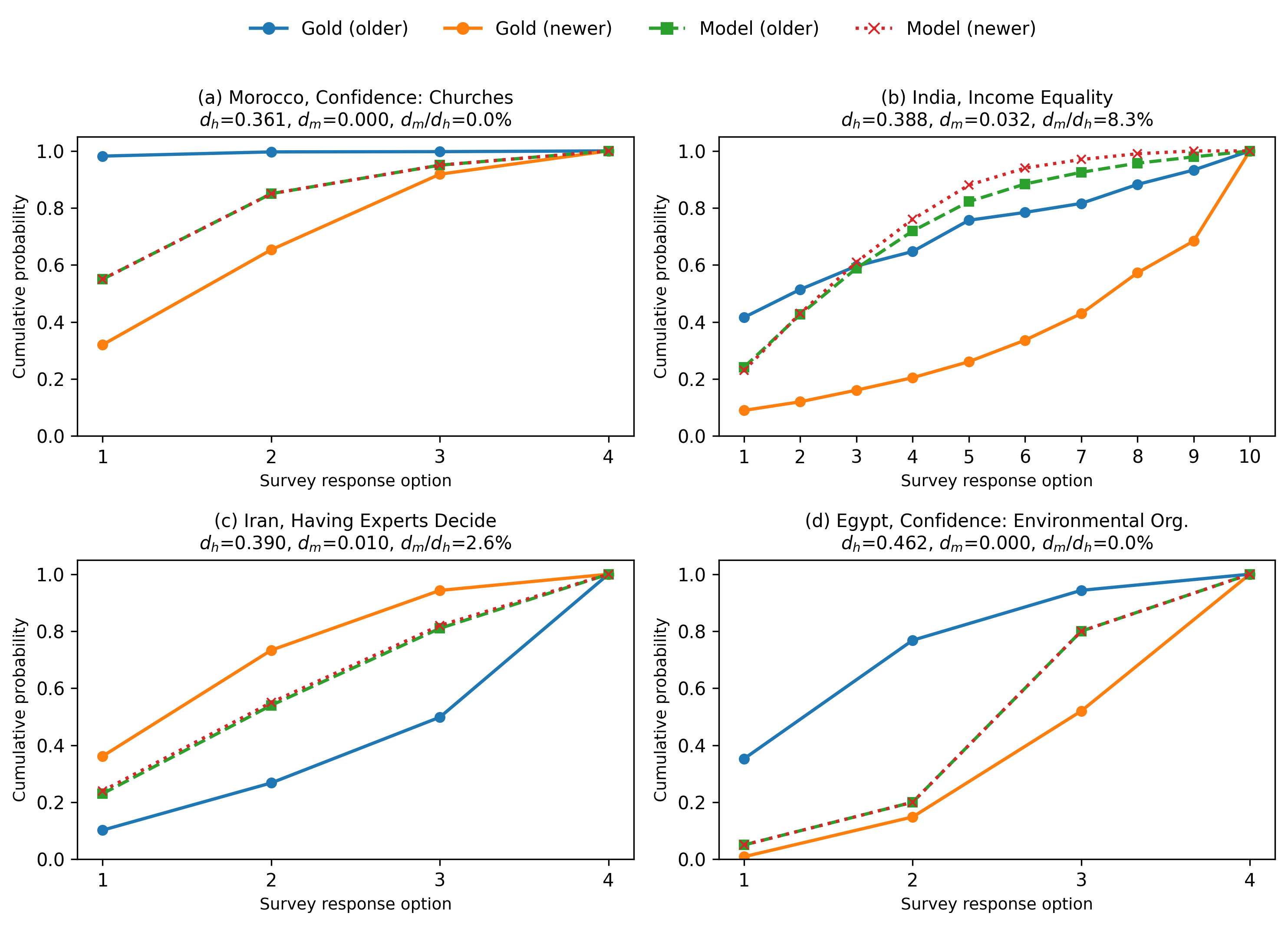}
    \caption{Selected examples of limited model
    change for Llama-3.3-70B despite substantial human shift:
    (a) Morocco, confidence in churches,
    (b) India, income equality,
    (c) Iran, having experts decide, and
    (d) Egypt, confidence in environmental
    organizations. Curves show cumulative
    response distributions. The annotations
    report human shift $d_h$, model shift $d_m$,
    and the magnitude ratio $d_m/d_h$ expressed
    as a percentage. Overlapping model curves
    indicate identical predictions for the
    older and newer periods.}
    \label{fig:rq2_examples}
\end{figure*}

\subsection{Illustrative Patterns of Predicted Change}
\label{app:rq2_arrows}

Figure~\ref{fig:rq2_arrows} illustrates the four outcomes discussed in
Section~\ref{sec:rq2}, using one survey question in four countries for
Llama-3.3-70B. For readability, each arrow shows the change in the
\emph{mean} response on the original answer scale between the older and
newer year. This is a simplified view: our direction measure in
Section~\ref{sec:rq2} compares changes across the full response
distribution, not only the mean.

\subsection{Examples of Limited Model Change}
\label{app:rq2_examples}

Figure~\ref{fig:rq2_examples} illustrates the
behavior examined in Section~\ref{sec:rq2}
using four selected country-question pairs.
Each panel shows the older and newer gold
distributions alongside the corresponding
model predictions, plotted as cumulative
response proportions.

Across these examples, the gold distributions
differ substantially, while the model predictions
overlap or remain close together. Morocco and
Egypt illustrate identical predictions for the
two periods, while India and Iran illustrate small
model changes relative to the human shift.
These examples make the limited model change
visible at the level of the response distributions.

\section{RF3: Default Period and Temporal Cues}
\subsection{Default Period of Undated Predictions}
\label{app:rq3_full}

Table~\ref{tab:rq3_default} reports the default survey year of undated predictions. Appendix~\ref{app:rq3_coverage} compares these rates with the rates expected if a year were chosen at random.

\begin{table}[!t]
\centering
\small
\setlength{\tabcolsep}{5pt}
\begin{tabular}{@{}lrrr@{}}
\toprule
\textbf{Model} & \textbf{Oldest} & \textbf{Interm.} & \textbf{Newest} \\
\midrule
Llama-3.3-70B  & 37.6 & 16.5 & 45.9 \\
GPT-4o         & 36.6 & 16.1 & 47.3 \\
GPT-5.4        & 37.0 & 17.1 & 45.9 \\
Qwen3.5-35B    & 38.3 & 15.5 & 46.2 \\
Qwen3.8-Flash  & 36.5 & 15.8 & 47.7 \\
DeepSeek-V3    & 37.7 & 16.4 & 45.8 \\
\bottomrule
\end{tabular}
\caption{Default survey year of undated predictions: the newest year is the most common, but earlier years account for more than half of items in every model. Cells are the \% of items whose undated prediction is closest to the oldest, an intermediate, or the newest available year.}
\label{tab:rq3_default}
\end{table}
\begin{table}[!t]
\centering
\small
\setlength{\tabcolsep}{4pt}
\renewcommand{\arraystretch}{1.1}
\begin{tabular}{llrr}
\toprule
\textbf{Available} & \textbf{Default} &
\textbf{Observed} & \textbf{Equal-choice} \\
\textbf{years} & \textbf{survey year} & & \\
\midrule
Exactly two
 & Oldest       & 45.36\% & 50.00\% \\
 & Newest       & 54.64\% & 50.00\% \\
\midrule
Three or more
 & Oldest       & 29.10\% & 30.94\% \\
 & Intermediate & 34.57\% & 38.12\% \\
 & Newest       & 36.33\% & 30.94\% \\
\bottomrule
\end{tabular}
\caption{Default survey year by survey coverage (Llama-3.3-70B, WVS and ISSP pooled): the newest year is selected only slightly more often than under equal choice. Equal-choice percentages are computed per item and averaged within groups. Items with two years have no intermediate year.}
\label{tab:rq3_coverage}
\end{table}

\subsection{Equal-Choice Comparison}
\label{app:rq3_coverage}

The number of available survey years differs
across country-question items. To help interpret
which years the undated predictions most closely
resemble, we introduce an \emph{equal-choice
baseline}. This baseline describes the expected
matching percentages if one available survey year
were selected at random for each item, with every
year having an equal chance.

For example, an item with two survey years has
a 50\% chance of selecting the oldest year and
a 50\% chance of selecting the newest. An item
with four years has a 25\% chance for each year:
25\% for the oldest, 25\% for the newest, and
50\% for the two intermediate years combined.

We calculate these probabilities separately
for each item and average them across items
to obtain the baseline percentages. Comparing
the observed matching percentages with this
baseline shows whether a category is matched
more or less often than under random year
selection.

Table~\ref{tab:rq3_coverage} reports this comparison
for Llama-3.3-70B, separately for items with exactly
two available survey years and those with three or more.

\subsection{Overall Effect of Naming a Year}
\label{app:rq3_overall}
Table \ref{tab:rq3} portrays the overall effect of naming a year.
\begin{table*}[!t]
\centering
\small
\setlength{\tabcolsep}{4pt}
\begin{tabular}{@{}lrrrrrrrrrr@{}}
\toprule
& \multicolumn{5}{c}{\textbf{Naming the older year}}
& \multicolumn{5}{c}{\textbf{Naming the newer year}} \\
\cmidrule(lr){2-6}\cmidrule(lr){7-11}
\textbf{Model} & Benefit & Helped & Unch. & Hurt & Ident.
& Benefit & Helped & Unch. & Hurt & Ident. \\
\midrule
Llama-3.3-70B  & $-0.78$ & 34.4 & 23.9 & 41.7 & 22.7 & $-0.69$ & 33.3 & 26.0 & 40.7 & 24.6 \\
GPT-4o         & $+0.02$ & 31.9 & 36.4 & 31.6 & 34.3 & $-0.17$ & 22.8 & 51.6 & 25.6 & 50.0 \\
GPT-5.4        & $-0.02$ & 44.2 & 12.1 & 43.7 & 10.6 & $-0.10$ & 41.7 & 18.0 & 40.3 & 16.5 \\
Qwen3.5-35B    & $-0.39$ & 30.3 & 38.6 & 31.0 & 37.9 & $-0.52$ & 26.2 & 42.8 & 31.0 & 41.9 \\
Qwen3.8-Flash  & $-0.09$ & 31.8 & 37.4 & 30.8 & 36.5 & $-0.36$ & 22.3 & 50.8 & 26.9 & 50.0 \\
DeepSeek-V3    & $-0.25$ & 26.5 & 44.2 & 29.3 & 42.6 & $-0.31$ & 20.4 & 54.9 & 24.7 & 53.7 \\
\bottomrule
\end{tabular}
\caption{Overall effect of naming a year: it helps and hurts similar shares of items, so the mean benefit ($\times 100$) is near zero for every model. Helped, Unch.\ (unchanged), and Hurt are \% of items and sum to 100; Ident.\ is the \% of items whose dated prediction equals the undated one.}
\label{tab:rq3}
\end{table*}

\subsection{Effect of Naming a Year by Human Shift}
\label{app:rq2_outcomes}

\begin{figure}[!htbp]
\centering
\includegraphics[width=0.8\columnwidth]{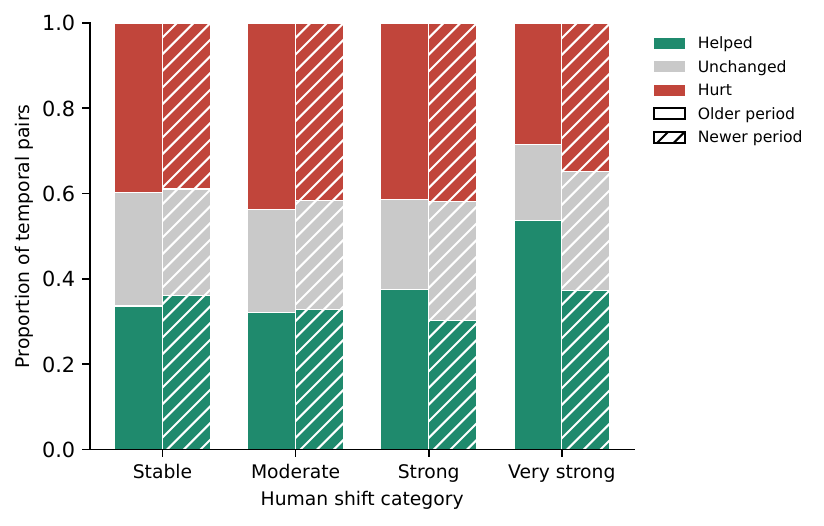}
\caption{Naming outcomes by human shift (Llama-3.3-70B): for very strong shifts, naming either period helps more often than it hurts.}
\label{fig:rq3_outcomes}
\end{figure}

Table~\ref{tab:rq3_by_shift} reports mean benefit by human shift for
all six models, as plotted in Figure~\ref{fig:rq3}. Bootstrap 95\% intervals over temporal pairs (1,000 resamples) confirm
these patterns. For stable and moderate shifts, the intervals for naming
either year lie below zero in all six models. For very strong shifts,
the interval for naming the older year lies above zero in all six
models, while the interval for naming the newer year includes zero in
five models and lies below zero for Qwen3.8-Flash. This benefit (about 0.03 in $W$) is small in absolute terms; we read it as a consistent direction rather than a large improvement. For Llama-3.3-70B,
Table~\ref{tab:rq2_outcomes} and Figure~\ref{fig:rq3_outcomes} also
report the share of items helped, unchanged, and hurt, using the
definitions in Section~\ref{sec:rq3}. Table~\ref{tab:naming_very_strong}
reports the shares helped and hurt for very strong shifts in all six
models.

\begin{table*}[!t]
\centering
\small
\setlength{\tabcolsep}{4pt}
\begin{tabular}{@{}lrrrrrrrr@{}}
\toprule
& \multicolumn{4}{c}{\textbf{Naming the older year}}
& \multicolumn{4}{c}{\textbf{Naming the newer year}} \\
\cmidrule(lr){2-5}\cmidrule(lr){6-9}
\textbf{Model} & Stab. & Mod. & Str. & V.\,str. & Stab. & Mod. & Str. & V.\,str. \\
\midrule
Llama-3.3-70B  & $-0.81$ & $-1.17$ & $-0.52$ & $+3.28$ & $-0.41$ & $-0.76$ & $-0.99$ & $-0.19$ \\
GPT-4o         & $-0.55$ & $-0.25$ & $+0.72$ & $+3.47$ & $-0.31$ & $-0.15$ & $-0.06$ & $-0.12$ \\
GPT-5.4        & $-0.43$ & $-0.37$ & $+0.79$ & $+3.62$ & $-0.15$ & $-0.15$ & $0.00$  & $+0.33$ \\
Qwen3.5-35B    & $-1.03$ & $-0.63$ & $+0.26$ & $+3.20$ & $-0.56$ & $-0.50$ & $-0.66$ & $-0.03$ \\
Qwen3.8-Flash  & $-0.63$ & $-0.43$ & $+0.61$ & $+4.55$ & $-0.36$ & $-0.36$ & $-0.30$ & $-0.49$ \\
DeepSeek-V3    & $-0.76$ & $-0.50$ & $+0.32$ & $+2.96$ & $-0.31$ & $-0.34$ & $-0.32$ & $+0.04$ \\
\bottomrule
\end{tabular}
\caption{Mean benefit ($\times 100$) by human shift: naming the older year helps after very strong shifts in every model, while naming the newer year helps little at any level. Plotted in Figure~\ref{fig:rq3}.}
\label{tab:rq3_by_shift}
\end{table*}

\begin{table}[!htbp]
\centering
\small
\setlength{\tabcolsep}{2pt}
\renewcommand{\arraystretch}{1.1}
\begin{tabular}{@{}llrrrr@{}}
\toprule
\textbf{Human} & \textbf{Period} & \textbf{Mean} &
\textbf{Helped} & \textbf{Unchanged} & \textbf{Hurt} \\
\textbf{shift} & & \textbf{benefit} & & & \\
\midrule
Stable
 & Older & $-0.81$ & 33.7\% & 26.6\% & 39.7\% \\
 & Newer & $-0.41$ & 36.1\% & 25.0\% & 38.9\% \\
\addlinespace
Moderate
 & Older & $-1.17$ & 32.2\% & 24.2\% & 43.6\% \\
 & Newer & $-0.76$ & 32.8\% & 25.6\% & 41.6\% \\
\addlinespace
Strong
 & Older & $-0.52$ & 37.6\% & 21.1\% & 41.4\% \\
 & Newer & $-0.99$ & 30.2\% & 27.9\% & 41.9\% \\
\addlinespace
Very strong
 & Older & $+3.28$ & 53.7\% & 17.7\% & 28.6\% \\
 & Newer & $-0.19$ & 37.3\% & 28.0\% & 34.8\% \\
\bottomrule
\end{tabular}
\caption{Naming outcomes by human shift (Llama-3.3-70B): naming either year helps more often than it hurts only for very strong shifts. Mean benefit is $\times 100$; helped, unchanged, and hurt mean positive, zero, and negative benefit. Percentages may not sum to 100 due to rounding.}
\label{tab:rq2_outcomes}
\end{table}

\begin{table}[!htbp]
\centering
\small
\setlength{\tabcolsep}{4pt}
\begin{tabular}{@{}lrrrr@{}}
\toprule
& \multicolumn{2}{c}{\textbf{Older year}} & \multicolumn{2}{c}{\textbf{Newer year}} \\
\cmidrule(lr){2-3}\cmidrule(lr){4-5}
\textbf{Model} & Helped & Hurt & Helped & Hurt \\
\midrule
Llama-3.3-70B  & 53.7 & 28.6 & 37.3 & 34.8 \\
GPT-4o         & 52.5 & 21.7 & 23.2 & 25.8 \\
GPT-5.4        & 63.7 & 29.1 & 47.9 & 39.7 \\
Qwen3.5-35B    & 49.0 & 21.7 & 30.3 & 30.3 \\
Qwen3.8-Flash  & 54.4 & 19.8 & 20.1 & 31.2 \\
DeepSeek-V3    & 53.2 & 20.5 & 25.7 & 26.0 \\
\bottomrule
\end{tabular}
\caption{Items helped and hurt by naming each period, very strong shifts (\%): naming the older year helps more often than it hurts in every model.}
\label{tab:naming_very_strong}
\end{table}

\section{Robustness Analyses}
\label{app:robustness}
This section tests whether our findings depend on the human-shift thresholds (Appendix~\ref{sec:threshold_robustness}), the knowledge-filter retention level (Appendix~\ref{app:gate_retention}), repeated runs and prompt instructions (Appendix~\ref{app:prompt_robustness}), and survey sampling noise (Appendix~\ref{app:sampling_noise}).

\subsection{Robustness to Human-Shift Thresholds}
\label{sec:threshold_robustness}

The four human-shift categories (Table~\ref{tab:shift_classification_counts})
depend on three boundaries. 

To check that our conclusions do not depend
on where these boundaries are drawn, we repeat the main analyses for all
six models under a tighter and a looser set of boundaries
(Table~\ref{tab:threshold_sets}). 
\begin{table}[!htbp]
\centering
\small
\setlength{\tabcolsep}{4pt}
\begin{tabular}{@{}lcccc@{}}
\toprule
\textbf{Set} & \textbf{Stable} & \textbf{Moderate} & \textbf{Strong} & \textbf{Very strong} \\
\midrule
Tighter & $\leq 0.02$ & $0.02$--$0.075$ & $0.075$--$0.15$ & $\geq 0.15$ \\
Default & $\leq 0.03$ & $0.03$--$0.10$  & $0.10$--$0.20$  & $\geq 0.20$ \\
Looser  & $\leq 0.04$ & $0.04$--$0.125$ & $0.125$--$0.25$ & $\geq 0.25$ \\
\bottomrule
\end{tabular}
\caption{Human-shift boundaries ($d_h$) used in the robustness check.
The default set is the one used in the main paper.}
\label{tab:threshold_sets}
\end{table}
Only the category assignment changes, and
all predictions and distances stay the same. For stable pairs, the
boundary for keeping a prediction steady moves with the stable boundary.

Tables~\ref{tab:threshold_rq1}--\ref{tab:threshold_stable} report the
results. 

\begin{table*}[!t]
\centering
\small
\setlength{\tabcolsep}{4pt}
\begin{tabular}{@{}llrrrrrrrr@{}}
\toprule
& & \multicolumn{4}{c}{\textbf{Both periods accurate (\%)}}
& \multicolumn{4}{c}{\textbf{One period only (\%)}} \\
\cmidrule(lr){3-6}\cmidrule(lr){7-10}
\textbf{Model} & \textbf{Set} & Stab. & Mod. & Str. & V.\,str. & Stab. & Mod. & Str. & V.\,str. \\
\midrule
Llama-3.3-70B & Tighter & 70.3 & 52.7 & 31.3 & 14.6 & 15.4 & 28.3 & 49.3 & 61.9 \\
 & Default & 65.1 & 47.8 & 22.8 & 11.5 & 19.0 & 33.2 & 55.9 & 63.7 \\
 & Looser & 62.2 & 42.6 & 18.0 & 9.2 & 20.9 & 38.2 & 60.1 & 58.8 \\
\addlinespace
GPT-4o & Tighter & 70.6 & 60.0 & 40.3 & 19.0 & 16.6 & 28.4 & 50.0 & 63.5 \\
 & Default & 66.9 & 55.9 & 31.1 & 13.7 & 20.2 & 33.4 & 56.5 & 67.2 \\
 & Looser & 66.4 & 50.9 & 25.0 & 6.7 & 21.5 & 38.2 & 62.0 & 64.2 \\
\addlinespace
GPT-5.4 & Tighter & 82.6 & 71.6 & 47.9 & 24.9 & 10.9 & 21.0 & 44.3 & 62.9 \\
 & Default & 80.3 & 66.4 & 37.0 & 19.5 & 12.3 & 26.4 & 54.3 & 66.1 \\
 & Looser & 78.4 & 61.1 & 30.7 & 11.2 & 14.6 & 31.5 & 59.5 & 67.3 \\
\addlinespace
Qwen3.5-35B & Tighter & 56.2 & 48.5 & 30.3 & 14.7 & 19.2 & 30.2 & 48.9 & 61.1 \\
 & Default & 54.3 & 44.4 & 23.4 & 10.4 & 21.5 & 35.0 & 55.4 & 63.2 \\
 & Looser & 53.5 & 39.9 & 19.7 & 6.2 & 23.4 & 39.3 & 60.0 & 53.8 \\
\addlinespace
Qwen3.8-Flash & Tighter & 71.5 & 60.4 & 40.9 & 20.5 & 14.2 & 27.9 & 47.0 & 63.0 \\
 & Default & 68.3 & 56.2 & 30.9 & 18.5 & 18.8 & 31.9 & 57.0 & 59.1 \\
 & Looser & 67.1 & 51.5 & 24.6 & 17.6 & 20.4 & 36.8 & 61.8 & 50.9 \\
\addlinespace
DeepSeek-V3 & Tighter & 62.8 & 54.2 & 36.9 & 16.1 & 16.9 & 29.2 & 48.5 & 66.5 \\
 & Default & 60.5 & 50.7 & 27.8 & 9.7 & 20.5 & 33.7 & 57.0 & 69.5 \\
 & Looser & 60.2 & 45.7 & 22.4 & 6.1 & 21.9 & 38.6 & 63.2 & 61.4 \\
\bottomrule
\end{tabular}
\caption{RF1 under alternative shift boundaries: for every model and boundary set, accuracy on both periods falls as human shift grows. One period only is older only plus newer only.}
\label{tab:threshold_rq1}
\end{table*}

\begin{table*}[!t]
\centering
\small
\setlength{\tabcolsep}{5pt}
\begin{tabular}{@{}llrrrrrr@{}}
\toprule
& & \multicolumn{3}{c}{\textbf{Separated (\%)}}
& \multicolumn{3}{c}{\textbf{Same period (\%)}} \\
\cmidrule(lr){3-5}\cmidrule(lr){6-8}
\textbf{Model} & \textbf{Set} & Mod. & Str. & V.\,str. & Mod. & Str. & V.\,str. \\
\midrule
Llama-3.3-70B & Tighter & 8.3 & 11.5 & 22.3 & 85.9 & 82.8 & 72.2 \\
 & Default & 8.8 & 15.2 & 26.7 & 85.5 & 79.1 & 68.9 \\
 & Looser & 9.7 & 17.8 & 29.8 & 84.5 & 76.6 & 66.4 \\
\addlinespace
GPT-4o & Tighter & 7.1 & 12.7 & 21.1 & 88.9 & 83.1 & 75.0 \\
 & Default & 8.7 & 15.0 & 23.9 & 87.2 & 81.0 & 73.9 \\
 & Looser & 9.7 & 18.3 & 25.0 & 86.3 & 77.5 & 73.3 \\
\addlinespace
GPT-5.4 & Tighter & 10.4 & 15.8 & 26.7 & 82.9 & 79.1 & 68.8 \\
 & Default & 11.8 & 19.6 & 29.8 & 82.0 & 75.5 & 67.1 \\
 & Looser & 13.0 & 23.9 & 24.3 & 81.0 & 71.2 & 73.8 \\
\addlinespace
Qwen3.5-35B & Tighter & 7.2 & 11.1 & 17.6 & 86.9 & 84.0 & 78.1 \\
 & Default & 8.5 & 13.1 & 19.6 & 85.9 & 81.8 & 77.7 \\
 & Looser & 9.5 & 15.3 & 15.2 & 84.9 & 80.3 & 83.4 \\
\addlinespace
Qwen3.8-Flash & Tighter & 8.7 & 13.2 & 22.9 & 86.2 & 81.6 & 74.2 \\
 & Default & 10.0 & 16.0 & 27.9 & 84.9 & 79.7 & 69.5 \\
 & Looser & 10.9 & 19.1 & 30.6 & 83.8 & 77.3 & 66.7 \\
\addlinespace
DeepSeek-V3 & Tighter & 6.6 & 9.1 & 17.8 & 89.4 & 87.5 & 79.4 \\
 & Default & 7.4 & 11.3 & 19.0 & 88.8 & 85.4 & 78.9 \\
 & Looser & 8.1 & 14.4 & 20.5 & 88.2 & 82.8 & 76.5 \\
\bottomrule
\end{tabular}
\caption{RF1 alignment under alternative shift boundaries: separated pairs become more common as human shift grows, but most pairs stay aligned with a single period under every set.}
\label{tab:threshold_alignment}
\end{table*}

\begin{table*}[!t]
\centering
\small
\setlength{\tabcolsep}{4pt}
\begin{tabular}{@{}llrrrrrrrrr@{}}
\toprule
& & \multicolumn{3}{c}{\textbf{Zero model shift (\%)}}
& \multicolumn{3}{c}{\textbf{Median ratio $r$}}
& \multicolumn{3}{c}{\textbf{Median cosine}} \\
\cmidrule(lr){3-5}\cmidrule(lr){6-8}\cmidrule(lr){9-11}
\textbf{Model} & \textbf{Set} & Mod. & Str. & V.\,str. & Mod. & Str. & V.\,str. & Mod. & Str. & V.\,str. \\
\midrule
Llama-3.3-70B & Tighter & 37.1 & 29.5 & 20.7 & 1.07 & 0.55 & 0.45 & 0.13 & 0.63 & 0.88 \\
 & Default & 35.8 & 25.2 & 17.4 & 0.84 & 0.50 & 0.39 & 0.22 & 0.76 & 0.88 \\
 & Looser & 33.1 & 23.7 & 18.3 & 0.73 & 0.48 & 0.37 & 0.36 & 0.84 & 0.90 \\
\addlinespace
GPT-4o & Tighter & 46.1 & 37.6 & 27.7 & 0.99 & 0.49 & 0.40 & 0.37 & 0.77 & 0.91 \\
 & Default & 44.2 & 32.3 & 28.3 & 0.80 & 0.44 & 0.36 & 0.50 & 0.84 & 0.91 \\
 & Looser & 42.4 & 30.0 & 25.8 & 0.67 & 0.40 & 0.35 & 0.62 & 0.90 & 0.90 \\
\addlinespace
GPT-5.4 & Tighter & 16.4 & 13.7 & 11.5 & 0.73 & 0.42 & 0.32 & 0.30 & 0.82 & 0.91 \\
 & Default & 15.8 & 11.8 & 10.3 & 0.60 & 0.38 & 0.26 & 0.45 & 0.88 & 0.92 \\
 & Looser & 15.2 & 10.8 & 11.2 & 0.51 & 0.37 & 0.20 & 0.62 & 0.91 & 0.91 \\
\addlinespace
Qwen3.5-35B & Tighter & 43.7 & 39.0 & 28.3 & 1.08 & 0.58 & 0.38 & 0.12 & 0.58 & 0.81 \\
 & Default & 42.7 & 33.5 & 28.2 & 0.85 & 0.47 & 0.30 & 0.21 & 0.71 & 0.86 \\
 & Looser & 41.0 & 30.6 & 26.2 & 0.73 & 0.42 & 0.25 & 0.30 & 0.78 & 0.79 \\
\addlinespace
Qwen3.8-Flash & Tighter & 45.4 & 37.9 & 31.0 & 1.13 & 0.55 & 0.43 & 0.27 & 0.71 & 0.88 \\
 & Default & 43.0 & 36.3 & 28.2 & 0.86 & 0.47 & 0.40 & 0.39 & 0.80 & 0.91 \\
 & Looser & 41.6 & 33.4 & 28.7 & 0.72 & 0.44 & 0.40 & 0.51 & 0.85 & 0.93 \\
\addlinespace
DeepSeek-V3 & Tighter & 57.4 & 51.2 & 36.0 & 1.23 & 0.56 & 0.36 & 0.34 & 0.73 & 0.83 \\
 & Default & 56.3 & 45.7 & 33.2 & 0.95 & 0.47 & 0.27 & 0.46 & 0.80 & 0.84 \\
 & Looser & 53.7 & 41.0 & 28.8 & 0.78 & 0.40 & 0.24 & 0.58 & 0.81 & 0.84 \\
\bottomrule
\end{tabular}
\caption{RF2 under alternative shift boundaries: for every model and boundary set, the median ratio falls and the median cosine rises as human shift grows. Ratio and cosine use pairs with nonzero model shift.}
\label{tab:threshold_rq2}
\end{table*}

\begin{table}[!htbp]
\centering
\small
\setlength{\tabcolsep}{4pt}
\begin{tabular}{@{}llrrr@{}}
\toprule
\textbf{Model} & \textbf{Set} & \textbf{Preserves} & \textbf{Small} & \textbf{Large} \\
\midrule
Llama-3.3-70B & Tighter & 59.6 & 32.4 & 8.0 \\
 & Default & 64.5 & 26.2 & 9.2 \\
 & Looser & 70.6 & 19.3 & 10.1 \\
\addlinespace
GPT-4o & Tighter & 63.5 & 32.9 & 3.6 \\
 & Default & 67.6 & 28.6 & 3.9 \\
 & Looser & 74.7 & 20.6 & 4.6 \\
\addlinespace
GPT-5.4 & Tighter & 54.7 & 43.3 & 2.0 \\
 & Default & 67.0 & 30.2 & 2.8 \\
 & Looser & 73.8 & 22.5 & 3.7 \\
\addlinespace
Qwen3.5-35B & Tighter & 64.7 & 26.6 & 8.7 \\
 & Default & 68.6 & 21.9 & 9.5 \\
 & Looser & 73.4 & 16.6 & 10.0 \\
\addlinespace
Qwen3.8-Flash & Tighter & 64.2 & 29.9 & 5.9 \\
 & Default & 67.2 & 24.5 & 8.3 \\
 & Looser & 72.4 & 18.7 & 9.0 \\
\addlinespace
DeepSeek-V3 & Tighter & 71.9 & 23.1 & 5.0 \\
 & Default & 70.1 & 22.9 & 7.0 \\
 & Looser & 78.2 & 14.3 & 7.5 \\
\bottomrule
\end{tabular}
\caption{Stable pairs under alternative shift boundaries (\%): most stable pairs keep a steady prediction under every set. Preserves uses each set's stable boundary; large drift is $d_m \geq 0.10$ throughout.}
\label{tab:threshold_stable}
\end{table}

Moving the boundaries changes the levels of each measure, since
each category then covers a different range of shifts, but not the
trends: under every boundary set and for every model, accuracy on both
periods falls as human shift grows, separated pairs become more common
but remain a minority, the median magnitude ratio falls while the median
cosine rises, and most stable pairs keep a steady prediction. Two details depend on the boundaries. Under the looser set, the share of
pairs with only one accurate period falls slightly from strong to very
strong shifts for four models, where the very strong category is small.
Under the tighter set, the median ratio for moderate shifts exceeds 1
for four models, so underestimation is clearest for strong and very
strong shifts.

\subsection{Robustness to Knowledge-Filter Retention}
\label{app:gate_retention}

The knowledge filter retains 75\% of question--country combinations
(Section~\ref{sec:4.1}). To check that our findings do not depend on
this choice, we repeat the main analyses for Llama-3.3-70B at 50\% and
100\% retention, using the same predictions and human-shift thresholds.
Only the set of retained combinations changes.
Tables~\ref{tab:gate_ret_rq12} and~\ref{tab:gate_ret_rq3} report the
results.

\begin{table*}[!t]
\centering
\small
\setlength{\tabcolsep}{4pt}
\begin{tabular}{@{}lrrrrrrrrrrrrr@{}}
\toprule
& \multicolumn{4}{c}{\textbf{Both accurate (\%)}}
& \multicolumn{3}{c}{\textbf{Separated (\%)}}
& \multicolumn{3}{c}{\textbf{Median ratio $r$}}
& \multicolumn{3}{c}{\textbf{Median cosine}} \\
\cmidrule(lr){2-5}\cmidrule(lr){6-8}\cmidrule(lr){9-11}\cmidrule(lr){12-14}
\textbf{Retained} & Stab. & Mod. & Str. & V.\,str.
& Mod. & Str. & V.\,str.
& Mod. & Str. & V.\,str.
& Mod. & Str. & V.\,str. \\
\midrule
50\%  & 80.7 & 59.9 & 27.0 & 13.9 & 9.8 & 15.9 & 26.0 & 0.84 & 0.50 & 0.39 & 0.23 & 0.75 & 0.89 \\
75\%  & 65.1 & 47.8 & 22.8 & 11.5 & 8.8 & 15.2 & 26.7 & 0.84 & 0.50 & 0.39 & 0.22 & 0.76 & 0.88 \\
100\% & 48.9 & 36.8 & 18.5 & 10.0 & 7.4 & 13.2 & 24.4 & 0.87 & 0.51 & 0.39 & 0.23 & 0.74 & 0.87 \\
\bottomrule
\end{tabular}
\caption{RF1 and RF2 at each knowledge-filter retention level (Llama-3.3-70B): all trends hold at every level. The 75\% row matches the main results. Accuracy uses $\tau=0.10$; ratio and cosine use pairs with nonzero model shift.}
\label{tab:gate_ret_rq12}
\end{table*}

\begin{table}[!htbp]
\centering
\small
\setlength{\tabcolsep}{4pt}
\begin{tabular}{@{}lrrrr@{}}
\toprule
& \textbf{Newest} & \multicolumn{2}{c}{\textbf{Benefit ($\times 100$)}} & \textbf{Preserves} \\
\cmidrule(lr){3-4}
\textbf{Retained} & \textbf{default (\%)} & Older & Newer & \textbf{(\%)} \\
\midrule
50\%  & 45.6 & $-1.31$ & $-0.99$ & 66.9 \\
75\%  & 45.9 & $-0.78$ & $-0.69$ & 64.5 \\
100\% & 46.9 & $-0.01$ & $-0.30$ & 63.7 \\
\bottomrule
\end{tabular}
\caption{RF3 and stable-pair results at each retention level (Llama-3.3-70B): the conclusions hold at every level. The 75\% row matches the main results. Newest default is the \% of combinations whose undated prediction is closest to the newest year; Preserves is the \% of stable pairs with $d_m \leq 0.03$.}
\label{tab:gate_ret_rq3}
\end{table}

Retaining fewer combinations generally raises accuracy, since the
remaining combinations are those for which the model shows the strongest
baseline knowledge. The trends, however, are unchanged. At every
retention level, accuracy on both periods falls as human shift grows,
separated pairs become more common but remain a minority, and the median
cosine rises. The median magnitude ratio stays below 1 and is lower for
strong and very strong shifts than for moderate ones. 
Undated predictions match the newest year in 46--47\% of cases,
specifying a year yields mean benefit near zero or slightly negative,
and most stable pairs keep a steady prediction (63.7--66.9\%).

\subsection{Robustness to Repeated Runs and Prompt Instructions}
\label{app:prompt_robustness}

We examine inference stability and sensitivity to elicitation instructions
using Llama-3.3-70B on 300 temporal pairs, sampled equally across the four
human-shift categories without applying the knowledge filter. First, we
repeat each original prompt five times. Predictions are identical to the
first run in 90.5\% of comparisons, with a mean distance of 0.0026,
indicating limited run-to-run variation. Second, we test two alternative
instructions that frame the task as estimating responses from a nationally
representative survey or from 100 adults. The survey questions and response
options remain unchanged, and within each temporal pair only the year
differs. Although these instructions affect individual predictions (mean
distances of 0.069 and 0.060 from the original instruction), the main
temporal patterns persist. Under all three instructions, both-period
accuracy declines with human shift, median magnitude ratios remain below 1
and are lower for strong and very strong shifts than for moderate shifts,
and directional alignment is highest for very strong shifts. Identical
predictions between periods also persist, although their frequency varies
across instructions.

\subsection{Survey Sampling Noise}
\label{app:sampling_noise}

Gold distributions are estimated from samples of respondents, so some
difference between years can arise from sampling alone. To estimate this
noise, we assume for each temporal pair that the older gold distribution
did not change, draw two independent samples of 500 respondents from it
(the minimum sample size in our benchmark), and compute the distance
between them, repeating this 200 times. Sampling alone produces a median
distance of 0.016, and the 95th percentile of these distances has a median
of 0.034 across pairs. Our stable threshold of 0.03 therefore corresponds
to differences that sampling alone can produce. All strong and very strong
shifts and 90.1\% of moderate shifts exceed the 95th percentile of their
own pair. Since most surveys include more than 500 respondents, these
estimates are conservative.
\makeatletter
\setlength{\@fptop}{0pt}\setlength{\@fpsep}{\textfloatsep}\setlength{\@fpbot}{0pt plus 1fil}
\setlength{\@dblfptop}{0pt}\setlength{\@dblfpsep}{\dbltextfloatsep}\setlength{\@dblfpbot}{0pt plus 1fil}
\makeatother
\end{document}